%% file: main.tex
\documentclass[11pt]{article}

\usepackage[T1]{fontenc}
\usepackage[utf8]{inputenc}
\usepackage{lmodern}
\usepackage[letterpaper,margin=1in]{geometry}
\usepackage{microtype}
\usepackage{xcolor}
\usepackage{graphicx}
\usepackage{booktabs}
\usepackage{multirow}
\usepackage{amsmath}
\usepackage{algorithm}
\usepackage{algpseudocode}
\usepackage{adjustbox}
\usepackage{array}
\usepackage[round,authoryear]{natbib}
\usepackage{url}
\usepackage{hyperref}

\input{math_commands.tex}

\definecolor{linkblue}{HTML}{1F4E79}
\hypersetup{
  hypertexnames=false,
  colorlinks=true,
  linkcolor=linkblue,
  citecolor=linkblue,
  urlcolor=linkblue,
  pdfauthor={Guoze Sun},
  pdftitle={Physics Transformer}
}
\newcommand{\methodname}[1]{\textsc{#1}}
\newcommand{\tableformat}{%
  \small
  \setlength{\tabcolsep}{4.5pt}%
  \renewcommand{\arraystretch}{1.08}%
}
\title{\textbf{Physics Transformer: Tailoring Transformer for General PDE Prediction}}
\author{
Guoze Sun\textsuperscript{1},
Rui Zhang\textsuperscript{1},
Jiankai Tang\textsuperscript{1},
Mengtao Yan\textsuperscript{1},
Runze Mao\textsuperscript{2},
Zhi X. Chen\textsuperscript{2},
Hao Sun\textsuperscript{1}
\\[0.5em]
\small
\textsuperscript{1}Gaoling School of Artificial Intelligence,
Renmin University of China, Beijing, China
\\
\small
\textsuperscript{2}School of Mechanics and Engineering Science,
Peking University, Beijing, China
\\[0.3em]
}
\date{}

\begin{document}

\maketitle
\begin{abstract}
Transformer architectures have attracted increasing attention for solving partial differential equations (PDEs), owing to their flexibility in handling irregular discretizations and their ability to capture long-range physical dependencies. However, unlike discrete language tokens or fixed-resolution image patches, observed physical fields are finite samples of underlying infinite-dimensional functions. Consequently, effectively applying Transformers to PDEs requires a tokenizer that respects the functional nature of physical fields and constructs physically expressive tokens from arbitrary discretizations.To this end, we propose \methodname{Physics Transformer}, a function-projection-based Transformer architecture for physical field prediction. Physics Transformer treats a physical field as a continuous function and partitions its discretization into locality-preserving spatial patches. Within each patch, it dynamically learns a set of adaptive local basis functions and projects the sampled field onto these bases to obtain compact physics tokens. The resulting tokens capture diverse latent physical states while preserving fine-scale spatial structures, enabling efficient global interaction through factorized attention across space and physical states. The projected representation further supports efficient decoding at arbitrary query locations. Extensive experiments on diverse benchmarks, ranging from two-dimensional PDE dynamics to industrial-scale three-dimensional CFD simulations, demonstrate that Physics Transformer accurately captures fine-grained physical structures and achieves state-of-the-art predictive performance. These results establish function projection as a practical and effective foundation for designing Transformer architectures for PDE solving.
\end{abstract}

\section{Introduction}
\label{sec:introduction}

Learning solution operators for PDE-governed systems is a central problem in
scientific machine learning. By directly approximating the mapping from input
conditions, coefficients, geometries, or source terms to physical fields,
neural operators provide fast surrogate models for fluid dynamics, wave
propagation, structural mechanics, and engineering design. Existing neural
operators have achieved remarkable progress on regular grids, where physical
fields can be naturally represented as dense tensors and processed by
convolutional, spectral, or operator-learning architectures~\citep{lu2021learning,li2021fourier,kovachki2023neural}.
However, many of these models become difficult to apply effectively to
irregular domains and complex geometries, where the underlying physical fields
are sampled on unstructured meshes, point clouds, or geometry-dependent
query locations rather than fixed Cartesian grids.

Several recent neural operators have been extended to irregular domains and
general geometries. Graph- or point-based neural operators naturally support
irregular inputs by operating on local neighborhoods or scattered points.
However, their local message passing mechanisms may require many layers to
propagate information across distant regions, making it difficult to efficiently capture long-range physical dependencies. Transformer-based neural operators, on the other hand, can model global physical interactions in a single attention layer by allowing each token to directly attend to all others, and are therefore well suited to general geometries.
However, the attention mechanism incurs quadratic complexity with respect
to the number of tokens, which severely challenges scalability on large-scale
discretizations.

Therefore, choosing appropriate tokens as the input of a Transformer operator
becomes crucial. In natural language processing and computer vision, tokens are naturally defined as words or image patches. For physical fields, however,
tokenization is more subtle. Mesh points should not be viewed as independent
pixels, but as discrete samples of an underlying continuous function. This
distinction is important because a physically meaningful token should not only
reduce the number of input elements, but also preserve the functional structure of the field. Existing approaches compress dense discretizations by projecting irregular points onto latent grids through local kernel integration, as in GINO, or by hierarchical graph aggregation followed by cross-attention, as in UPT. These approaches provide effective discretization-independent compression. However, their latent units remain generic grid features or learned latent slots, rather than explicit representations of the local functional modes of the field. Consequently, fine-scale variations within each local region must be implicitly folded into a single feature representation and may be lost under aggressive compression.

Transolver and its variants take an important step toward physics-aware
tokenization. Instead of attending on all mesh points level, Transolver
introduces Physics-Attention, which adaptively aggregates mesh points into
learnable slices and performs attention among the resulting physical slices~\citep{wu2024transolver}, leading to more meaningful physical representations. However, tokenization is performed globally over the entire domain. Although physically different subregions could be compressed into different slices, global aggregation could over-smooth local fine-scale structures. Further, despite in similar physical states, local fields could still evolve heterogeneously according to positions and surrounding influences, so explicitly modeling local interaction instead of global aggregation is essential to capture complex dynamics.

To address these challenges, we propose \methodname{Physics Transformer}, a Transformer architecture tailored to PDE prediction. Given an irregular discretization, \methodname{Physics Transformer} first orders the input points using Hilbert space-filling curve to partition into P contiguous patches, providing an efficient and robust locality-preserving decomposition. Within each patch, a globally shared basis generator dynamically constructs M input-adaptive basis functions, each representing a distinct latent physical state. The local physical field is then projected onto these basis functions to obtain M compact physics tokens. Since the basis generator and its ordered basis slots are shared across all patches, tokens with the same basis index are encouraged to encode semantically aligned physical states in different spatial regions. This yields a structured token tensor of size P×M, over which we perform factorized attention. Specifically, global attention is first applied across patches between tokens representing the same physical state, efficiently capturing long-range physical coupling. Local attention is subsequently applied across different physical states within each patch to model their local interactions and evolution. Compared with full self-attention over all PM tokens, the proposed factorization reduces the attention complexity while preserving both global physical communication and local state interaction. Finally, we decode these tokens back into physical fields using another set of adaptive basis functions.

We further demonstrate that the tokens learned by our model is a compact representation of the local fields rather than the input points. We come up with a mechanism that could query the value of arbitrary points using the cached tokens during inference on a subset of input mesh with the same accuracy, which could accelerate inference on industrial-scale meshes like DriverML.

We further provide a projected-operator interpretation of \methodname{Physics Transformer}.Inspired by Galerkin-type attention~\citep{cao2021choose}, the patch-wise encoder and decoder can be viewed as learnable test and trial spaces, while the factorized token attention learns a finite-dimensional operator between them.

We evaluate Physics Transformer on a diverse set of PDE tasks, including 2D regular-grid problems, irregular-mesh problems with complex physical phenomena, and large-scale 3D CFD datasets. Across these settings,
Physics Transformer achieves state-of-the-art accuracy with favorable efficiency,
demonstrating its ability to handle both regular and irregular domains while
preserving complex physical structures.

Our contributions are summarized as follows:
\begin{itemize}
    \item We propose Physics Transformer, a Transformer architecture tailored to
    physical field prediction on general geometries. The model reformulates Transformer-based operator learning as a problem of constructing meaningful physics tokens from continuous fields.

    \item We introduce patch-wise physics tokenization based on function projection which provides a more expressive representation of physical fields and factorized attention for efficient global interaction and local evolving.

    \item We propose a mechanism enabling Physics Transformer could decode at arbitrary points, facilitating efficient inference on industry scale tasks.

    \item We provide a projected-operator interpretation of Physics Transformer and
    demonstrate its effectiveness on 2D regular grids, irregular meshes, and
    large-scale 3D CFD datasets, where it achieves state-of-the-art performance with favorable efficiency.
\end{itemize}

\section{Related Work}
\label{sec:related_work}

\paragraph{Neural operators on regular and irregular domains.}
Neural operators learn mappings between function spaces for
PDE-governed systems. DeepONet represents solution operators through
a branch--trunk decomposition with learned coordinate-dependent basis
functions~\citep{lu2021learning}, while FNO performs global operator
learning using truncated Fourier modes~\citep{li2021fourier}. Although
effective on structured domains, their standard formulations are less
suited to unstructured meshes and geometry-dependent discretizations.
Graph neural operators and mesh-based simulators address this issue
through neighborhood message passing~\citep{li2020neural,pfaff2021learning}.
Geo-FNO learns a deformation from irregular physical domains to regular
computational domains~\citep{li2022fourier}, whereas GINO encodes
irregular inputs onto a latent grid, applies Fourier operators, and
decodes predictions at arbitrary query locations~\citep{li2023geometry}.
These extensions improve geometric flexibility, but local propagation
and intermediate grid representations can make it difficult to
simultaneously preserve fine-scale structures and capture long-range
physical coupling.

\paragraph{Patchification of irregular 3D discretizations.}
Unlike images, irregular meshes and point clouds lack a canonical
partition into spatially coherent and computationally balanced patches.
Point Transformer V3 voxelizes and serializes 3D points using
space-filling curves, including Z-order and Hilbert curves, and groups
contiguous points into fixed-size patches for efficient local
attention~\citep{wu2024pointtransformerv3}. SpiderSolver constructs
boundary-guided spiderweb-like subregions and combines coarse-grained
inter-patch attention with fine-grained interactions near domain
boundaries~\citep{qi2025spidersolver}. Erwin instead recursively
organizes irregular points with a ball tree, performing attention within
equal-cardinality partitions and progressively coarsening and refining
the hierarchy to enlarge the receptive field~\citep{zhdanov2025erwin}.
These approaches expose a fundamental trade-off among partitioning cost,
spatial locality, and cross-patch communication. In our method, Hilbert
traversal provides an efficient and locality-preserving partition of
arbitrary discretizations; importantly, the resulting patches serve as
local function domains on which adaptive physical basis functions are
constructed, rather than merely as windows for local attention.

\paragraph{Transformer-based neural operators and physical tokenization.}
Transformer-based operators use attention to model nonlocal interactions
on flexible discretizations. OFormer and GNOT combine self- and
cross-attention to learn operators and evaluate solutions at arbitrary
queries~\citep{li2022transformer,hao2023gnot}. UPT compresses simulation
inputs into a compact latent token space before propagation and
query-based decoding~\citep{alkin2024universal}, while GAOT combines
geometry-aware graph encoders and decoders with Transformer
processors~\citep{wen2025geometry}. A closely related direction is the
Transolver family. Transolver softly assigns mesh points to a small set
of physics-aware slices, performs attention among slice tokens, and
maps the updated representations back to the original
mesh~\citep{wu2024transolver}. Subsequent variants improve scalability,
geometric conditioning, or computational efficiency
~\citep{luo2025transolverpp,wu2026transolver3,
hu2025linearno,adams2025geotransolver}. Nevertheless, they largely retain
the global construction in which the entire domain is aggregated into a
shared set of latent physical states. Such global aggregation can obscure fine-grained physical structures and limit the modeling of localized interactions. Moreover, Transolver
reports that increasing the number of global slice tokens yields
diminishing returns and can even degrade performance
~\citep{wu2024transolver}. Therefore, simply increasing the slice count
does not fundamentally resolve the representational bottleneck introduced
by global compression. In contrast, our method learns adaptive basis functions independently within each spatial patch and
projects the local physical field onto these bases. This patch-wise
functional tokenization preserves heterogeneous local states while
allowing global physical coupling through attention over the resulting
tokens.

\section{Method}
\label{sec:method}

\begin{figure}[!t]
    \centering
    \includegraphics[width=1\linewidth]{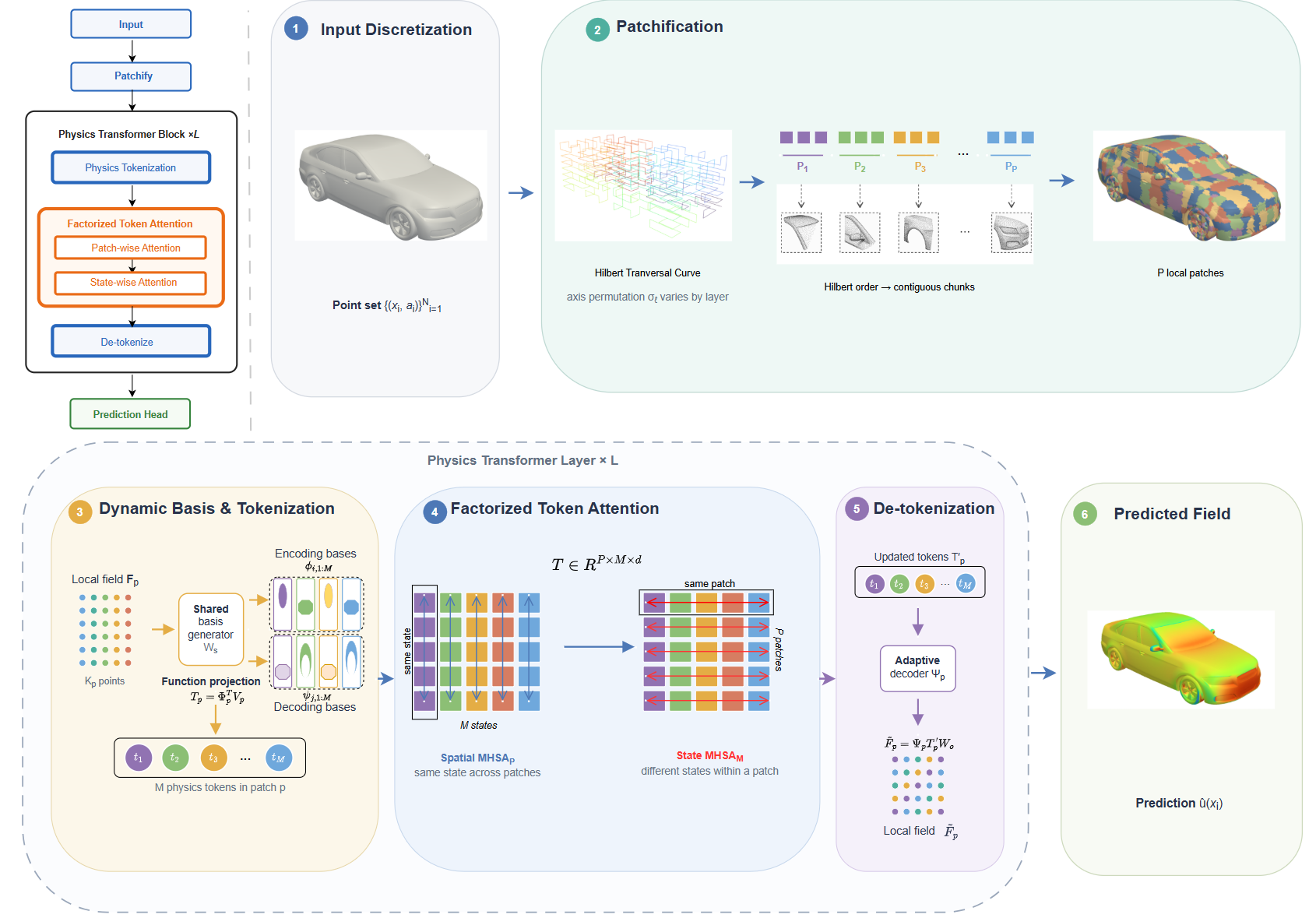}
    \caption{The main structure of Physics Transformer. A general overview is on top left. And a detailed demonstration is provided focusing on the Patchification and Physics Transformer block for concrete understanding.}
    \label{fig:model}
\end{figure}

We propose Physics Transformer, a Transformer-based neural solver for learning
solution maps on regular grids, irregular meshes, and point clouds. Given a
discretized physical domain
\(\mathcal{X}=\{\mathbf{x}_i\}_{i=1}^{N}\subset\mathbb{R}^{d}\), input features
\(\mathbf{a}_i\in\mathbb{R}^{c_{\rm in}}\), and target physical fields
\(\mathbf{u}_i\in\mathbb{R}^{c_{\rm out}}\), our goal is to approximate the
operator
\[
    \mathcal{G}: \{(\mathbf{x}_i,\mathbf{a}_i)\}_{i=1}^{N}
    \mapsto \{\mathbf{u}_i\}_{i=1}^{N}.
\]
Instead of applying self-attention directly to all discretization points,
Physics Transformer first partitions the domain into local patches, then compresses
each patch into a small set of learned \emph{physics tokens}. Global interaction
is performed over these compact physics tokens, and the updated tokens
are decoded back to point-level fields. This design avoids dense global attention
over all points while preserving local patch-conditioned physical representations.

\subsection{Patchification}
\label{sec:patchify}

We first partition the input discretization into \(P\) local patches
\(\{\mathcal{P}_p\}_{p=1}^{P}\). For regular Cartesian grids, we use
non-overlapping grid windows, following the standard patch partitioning strategy
in vision Transformers~\citep{dosovitskiy2021image,liu2021swin}. For irregular
2D/3D meshes or point clouds, we serialize points with a locality-preserving
space-filling curve and then split the ordered sequence into contiguous
patches~\citep{wu2024pointtransformerv3}.

Concretely, we normalize coordinates to \([0,1]^d\), quantize them onto a
discrete lattice with resolution \(2^L\), and compute a Hilbert serialization
code
\[
    h_i^{(\ell)}
    =
    \operatorname{Hilbert}_{L}
    \Big(
        \mathbf{Q}_{L}
        \big(
            \mathbf{P}_{\sigma_\ell}\,\bar{\mathbf{x}}_i
        \big)
    \Big),
    \qquad
    \bar{\mathbf{x}}_i\in[0,1]^d ,
\]
where \(\mathbf{Q}_{L}\) denotes coordinate quantization and
\(\mathbf{P}_{\sigma_\ell}\) is an axis permutation used at layer \(\ell\).
Sorting points by \(h_i^{(\ell)}\) gives a permutation \(\pi_\ell\). We then
define
\[
    \mathcal{P}_{p}^{(\ell)}
    =
    \left\{
        \pi_\ell(j)
        \ \middle|\
        (p-1)K_\ell < j \leq pK_\ell
    \right\},
    \qquad
    K_\ell=\left\lceil \frac{N}{P} \right\rceil .
\]
Changing \(\sigma_\ell\) across layers gives a simple Trans-Hilbert
patchification, which changes patch boundaries and allows points near a patch
boundary in one layer to interact with different neighbors in another layer.
In our implementation, we fix the number of patches \(P\), rather than fixing the
number of points per patch. This makes each patch correspond to a physically
comparable region of the domain under different sampling densities.

\subsection{Patch-wise Physics Tokenization}
\label{sec:physics_tokenization}

Let \(\mathbf{F}^{0}\in\mathbb{R}^{N\times d_h}\) denote the initial point
embedding:
\[
    \mathbf{f}_i^{0}
    =
    \phi_{\rm in}
    \left(
        \left[
            \mathbf{a}_i,\,
            \gamma(\mathbf{x}_i)
        \right]
    \right),
\]
where \(\gamma(\cdot)\) is a coordinate encoding. For the \(p\)-th patch, we
write
\[
    \mathbf{F}_{p}^{0}\in\mathbb{R}^{K_p\times d_h}
\]
for the features of points inside \(\mathcal{P}_p\), where \(K_p\) is the number
of points in the patch.

Each patch is compressed into \(M\) learned physics tokens through a function-projection process. For the \(p\)-th patch, the point-wise feature field
\(\mathbf{F}_{p}^{0}\in\mathbb{R}^{K_p\times d_h}\) is first mapped to token
projection scores:
\[
    \mathbf{S}_{p}
    =
    \mathbf{F}_{p}^{0}\mathbf{W}_{s},
    \qquad
    \mathbf{S}_{p}\in\mathbb{R}^{K_p\times M}.
\]
We then apply two nonlinear operators to obtain the tokenization basis functions
and the de-tokenization basis functions:
\[
    \boldsymbol{\Phi}_{p}
    =
    \mathcal{N}_{K_p}(\mathbf{S}_{p}),
    \qquad
    \boldsymbol{\Psi}_{p}
    =
    \mathcal{N}_{M}(\mathbf{S}_{p}),
\]
where \(\mathcal{N}_{K_p}=\operatorname{softmax}_{K_p}\) normalizes over the
point dimension, and \(\mathcal{N}_{M}=\operatorname{softmax}_{M}\) normalizes
over the token dimension. Thus,
\(\boldsymbol{\Phi}_{p}\in\mathbb{R}^{K_p\times M}\) defines \(M\) local learned
tokenization basis functions, while
\(\boldsymbol{\Psi}_{p}\in\mathbb{R}^{K_p\times M}\) defines the corresponding
de-tokenization basis functions for reconstructing point-wise fields from
physics tokens.

With point values
\[
    \mathbf{V}_{p}
    =
    \mathbf{F}_{p}^{0}\mathbf{W}_{v},
    \qquad
    \mathbf{V}_{p}\in\mathbb{R}^{K_p\times d_t},
\]
the \(p\)-th patch is encoded into \(M\) physics tokens:
\[
    \mathbf{T}_{p}
    =
    \mathbf{K}_{p}^{\top}\mathbf{V}_{p},
    \qquad
    \mathbf{T}_{p}\in\mathbb{R}^{M\times d_t}.
\]
Stacking all patches gives
\[
    \mathbf{T}
    =
    \{\mathbf{T}_{p,m}\}_{p=1,m=1}^{P,M}
    \in
    \mathbb{R}^{P\times M\times d_t}.
\]
Since the sprojection \(\mathbf{W}_{s}\) is shared across patches, the
same token index \(m\) represents a shared latent physics state in different
local regions. Thus, Physics Transformer localizes physics tokenization before global
interaction: each patch learns its own token representation, while token indices
remain aligned across the whole domain.

\subsection{Factorized Global Attention over Physics Tokens}
\label{sec:global_attention}

The token tensor \(\mathbf{T}\in\mathbb{R}^{P\times M\times d_t}\) contains
\(P\) spatial patches and \(M\) physics tokens per patch. We perform global
interaction in a factorized manner.

First, for each token index \(m\), we perform spatial self-attention over all
patches:
\[
    \mathbf{U}_{:,m}
    =
    \mathbf{T}_{:,m}
    +
    \operatorname{MHSA}_{P}
    \left(
        \operatorname{LN}(\mathbf{T}_{:,m});
        \mathbf{C}
    \right),
    \qquad
    m=1,\dots,M ,
\]
where \(\mathbf{C}=\{\mathbf{c}_{p}\}_{p=1}^{P}\) are patch centers,
\[
    \mathbf{c}_{p}
    =
    \frac{1}{K_p}\sum_{i\in\mathcal{P}_{p}}\mathbf{x}_i .
\]
The patch centers provide spatial information for long-range interaction among
the same physics token across different regions.

Second, for each patch \(p\), we perform token-wise self-attention across the
\(M\) physics tokens within the same patch:
\[
    \mathbf{Z}_{p,:}
    =
    \mathbf{U}_{p,:}
    +
    \operatorname{MHSA}_{M}
    \left(
        \operatorname{LN}(\mathbf{U}_{p,:})
    \right),
    \qquad
    p=1,\dots,P .
\]
Finally, a token-wise feed-forward network gives
\[
    \mathbf{T}_{p,:}^{\prime}
    =
    \mathbf{Z}_{p,:}
    +
    \operatorname{FFN}
    \left(
        \operatorname{LN}(\mathbf{Z}_{p,:})
    \right).
\]
The first stage captures global interactions among the same physics token across
the domain, while the second stage models interactions among different physics
tokens inside each local region. This factorization enables explicit global
communication without applying attention to all point tokens.

\subsection{De-tokenization and Output Prediction}
\label{sec:detokenization}

After global token mixing, the updated physics tokens
\[
    \mathbf{T}_{p}^{\prime}\in\mathbb{R}^{M\times d_t}
\]
are mapped back to point-level features through the de-tokenization basis family
\(\boldsymbol{\Psi}_{p}\). Recall that
\(\boldsymbol{\Psi}_{p}\) defines how each point in patch \(p\) reconstructs its
feature from the \(M\) updated physics tokens.

The de-tokenized point-level features are computed as
\[
    \overline{\mathbf{F}}_{p}
    =
    \boldsymbol{\Psi}_{p}
    \mathbf{T}_{p}^{\prime}
    \mathbf{W}_{o},
    \qquad
    \overline{\mathbf{F}}_{p}\in\mathbb{R}^{K_p\times d_h}.
\]
Therefore, the output at each point is reconstructed from the learned
patch-wise physics tokens.

The decoded features from all patches are restored to the original point
ordering and passed through an output head:
\[
    \widehat{\mathbf{u}}_i
    =
    \phi_{\rm out}
    \left(
        \overline{\mathbf{f}}_i
    \right),
    \qquad
    i=1,\dots,N .
\]
For time-dependent prediction, the predicted field can be used autoregressively
to produce multi-step rollout predictions.
\subsection{Arbitrary Query Decoding}
\label{sec:arbitrary_query}

A key property of Physics Transformer is that its prediction is not tied to the input
discretization points. In Physics Transformer, the learned physics tokens are obtained
through a function-projection process: local point-wise fields are projected into
compact patch-wise tokens, which serve as latent representations of local
physical fields. Therefore, for a given input instance, we first perform one
forward pass on a set of support points and cache the intermediate physics tokens
after global token interaction. Once these tokens are stored, predictions at new
spatial locations are produced only through de-tokenization from the cached
patch-wise physics tokens, without re-running the encoder or the global token
interaction.

Let
\[
\mathcal{S}=\{(\mathbf{x}_i,\mathbf{a}_i)\}_{i=1}^{N_s}
\]
be a support point set used to encode the input function, where \(N_s\) can be
much smaller than the full mesh size. Physics Transformer performs one encoding
forward pass on \(\mathcal{S}\), producing updated patch-wise physics tokens
\[
\mathbf{T}' =
E_{\theta}(\mathcal{S})
\in
\mathbb{R}^{P\times M\times d_t},
\]
where \(P\) is the number of patches and \(M\) is the number of physics tokens
per patch. During this forward pass, we also record the patch index \(p_i\) of
each support point \(\mathbf{x}_i\). The cached tokens \(\mathbf{T}'\) provide
an input-dependent latent representation of the solution field over the domain.

Given a new query point \(\mathbf{x}_q\), we do not re-run tokenization or global
token interaction. Instead, we find its nearest support point
\[
n(q)
=
\arg\min_{i\in\{1,\dots,N_s\}}
\left\|
\mathbf{x}_q-\mathbf{x}_i
\right\|_2 ,
\]
and use the patch index of this nearest support point to select the corresponding
cached physics-token bank:
\[
p(q)=p_{n(q)}.
\]
Then its de-tokenization weights are computed:
\[
\mathbf{q}_q
=
\operatorname{softmax}
\left(
\phi_q([\gamma(\mathbf{x}_q),\mathbf{a}_q])\mathbf{W}_q
\right)
\in
\mathbb{R}^{M},
\]
where \(\gamma(\mathbf{x}_q)\) is the coordinate encoding and \(\mathbf{a}_q\)
denotes optional query features when available. The prediction at
\(\mathbf{x}_q\) is then obtained by de-tokenizing from the selected cached
patch-wise physics tokens:
\[
\widehat{\mathbf{u}}(\mathbf{x}_q)
=
\phi_{\rm out}
\left(
\mathbf{q}_q^{\top}
\mathbf{T}'_{p(q)}
\mathbf{W}_o
\right).
\]

For a query set
\[
\mathcal{Q}=\{\mathbf{x}_q\}_{q=1}^{N_q},
\]
the same cached tokens can be decoded repeatedly:
\[
\widehat{\mathcal{U}}_{\mathcal{Q}}
=
D_{\theta}(\mathcal{Q},\mathbf{T}',\mathcal{S}).
\]

Therefore, Physics Transformer follows the neural-operator paradigm: the support
points are used to encode the input function into patch-wise physics tokens,
while arbitrary query locations are evaluated by selecting the nearest
support-associated token bank and computing query-specific de-tokenization
weights. Changing the query resolution only requires nearest-neighbor lookup and
lightweight de-tokenization, instead of recomputing the encoder or global token
interaction.

\subsection{Complexity}
\label{sec:method_complexity}

We analyze the token-mixing cost of one Physics Transformer block. Let \(N=PK\), where
\(P\) is the number of patches and \(K\) is the average number of points per
patch. Let \(M\) be the number of physics tokens per patch and \(d_t\) be the
token dimension.

Patch-wise physics tokenization and de-tokenization cost
\[
    O(NMd_t).
\]
The global attention over the same token index is performed over \(P\) patches
for each of the \(M\) token indices, giving
\[
    O(MP^2d_t).
\]
The token-wise attention inside each patch costs
\[
    O(PM^2d_t).
\]
Therefore, the total token-mixing complexity is
\[
    O(NMd_t + MP^2d_t + PM^2d_t).
\]
In contrast, applying global self-attention to all discretization points requires
\[
    O(N^2d_h)
\]
time and \(O(N^2)\) attention memory. Since \(M\ll K\) and \(P\ll N\) in
large-scale physical fields, Physics Transformer substantially reduces the cost of
global interaction while retaining explicit long-range communication through
patch-wise physics tokens.

\section{Analysis}
\label{sec:analysis}

We provide an input-adaptive projected-operator interpretation of
\methodname{Physics Transformer}. The key observation is
that \methodname{Physics Transformer} never constructs a dense global interaction matrix over the
\(N\) discretization points. Instead, it projects point-level features onto
\(PM\) patch-wise physics tokens, performs nonlocal interaction in the
compressed token space, and reconstructs the point-level field. This leads to
an expressive projected neural operator without quadratic point-wise attention.

\subsection{Connection to Galerkin-Type Attention}
\label{sec:galerkin_connection}

Our interpretation is inspired by the Galerkin
Transformer~\citep{cao2021choose}, which connects attention-based neural
operators with classical Petrov--Galerkin projection. Standard self-attention
performs global information exchange through
\[
    \operatorname{Attn}(\mathbf{Q},\mathbf{K},\mathbf{V})
    =
    \operatorname{softmax}
    \left(
        \frac{\mathbf{Q}\mathbf{K}^{\top}}{\sqrt{d_t}}
    \right)
    \mathbf{V},
\]
which explicitly forms an \(N\times N\) point-wise interaction matrix.
Galerkin-type attention removes the softmax normalization and reassociates the
matrix products as
\[
    (\mathbf{Q}\mathbf{K}^{\top})\mathbf{V}
    =
    \mathbf{Q}
    \left(
        \mathbf{K}^{\top}\mathbf{V}
    \right).
\]
When the feature dimension is fixed, this formulation has linear complexity
with respect to \(N\). It can also be interpreted as a learnable projection:
\(\mathbf{K}^{\top}\) tests the input field against learned test functions,
while \(\mathbf{Q}\) reconstructs the output in a learned trial space. The
attention layer therefore acts as a finite-dimensional projected operator
analogous to a Petrov--Galerkin method.

\methodname{Physics Transformer} follows this projection principle but constructs its test and
trial functions locally. For each of the \(P\) patches, it learns \(M\)
tokenization basis functions \(\boldsymbol{\Phi}_p\) and \(M\)
de-tokenization basis functions \(\boldsymbol{\Psi}_p\). Because basis
functions associated with different patches have disjoint supports,
\methodname{Physics Transformer} has access to as many as \(PM\) patch-wise basis functions,
rather than only \(M\) globally supported functions. The projected
coefficients are then globally coupled by factorized token attention. Unlike
a classical projection method, both the basis functions and the coefficient
map are input-dependent, making \methodname{Physics Transformer} an adaptive nonlinear projected
operator.

\subsection{Patch-Wise Projected Operator}
\label{sec:projected_operator}

Consider one Physics Transformer layer and suppress the layer index \(\ell\) for
clarity. Let
\[
    \mathcal{X}
    =
    \{\mathbf{x}_i\}_{i=1}^{N}
\]
be partitioned into
\[
    \{\mathcal{P}_p\}_{p=1}^{P},
    \qquad
    K_p = |\mathcal{P}_p|,
    \qquad
    \sum_{p=1}^{P}K_p=N.
\]
For patch \(\mathcal{P}_p\), let
\[
    \mathbf{F}_p^{0}\in\mathbb{R}^{K_p\times d_h}
\]
denote its point-level features. As defined in
Section~\ref{sec:physics_tokenization}, the projection scores and value
features are
\[
    \mathbf{S}_p
    =
    \mathbf{F}_p^{0}\mathbf{W}_s
    \in\mathbb{R}^{K_p\times M},
    \qquad
    \mathbf{V}_p
    =
    \mathbf{F}_p^{0}\mathbf{W}_v
    \in\mathbb{R}^{K_p\times d_t}.
\]
The tokenization and de-tokenization basis matrices are
\[
    \boldsymbol{\Phi}_p
    =
    \mathcal{N}_{K_p}(\mathbf{S}_p),
    \qquad
    \boldsymbol{\Psi}_p
    =
    \mathcal{N}_{M}(\mathbf{S}_p),
\]
where
\[
    \boldsymbol{\Phi}_p,\boldsymbol{\Psi}_p
    \in\mathbb{R}^{K_p\times M}.
\]
The patch-wise projection is therefore
\[
    \mathbf{T}_p
    =
    \boldsymbol{\Phi}_p^{\top}\mathbf{V}_p
    \in\mathbb{R}^{M\times d_t}.
\]
Stacking the tokens across all patches gives
\[
    \mathbf{T}
    =
    \{\mathbf{T}_{p,m}\}_{p=1,m=1}^{P,M}
    \in\mathbb{R}^{P\times M\times d_t}.
\]
The factorized global attention module updates these coefficients as
\[
    \mathbf{T}^{\prime}
    =
    \mathcal{H}_{\theta}(\mathbf{T}),
\]
where \(\mathcal{H}_{\theta}\) consists of attention along the patch
dimension, attention along the physics-token dimension, and the token-wise
feed-forward network. Finally, point-level features are reconstructed by
\[
    \overline{\mathbf{F}}_p
    =
    \boldsymbol{\Psi}_p
    \mathbf{T}_p^{\prime}
    \mathbf{W}_o
    \in\mathbb{R}^{K_p\times d_h}.
\]

To express this operation globally, reorder the points by their patch
membership and define the block-local matrices
\[
    \boldsymbol{\Phi}
    =
    \operatorname{blkdiag}
    \left(
        \boldsymbol{\Phi}_1,\ldots,\boldsymbol{\Phi}_P
    \right),
    \qquad
    \boldsymbol{\Psi}
    =
    \operatorname{blkdiag}
    \left(
        \boldsymbol{\Psi}_1,\ldots,\boldsymbol{\Psi}_P
    \right),
\]
where
\[
    \boldsymbol{\Phi},\boldsymbol{\Psi}
    \in\mathbb{R}^{N\times PM}.
\]
Similarly, let
\[
    \mathbf{V}
    =
    \begin{bmatrix}
        \mathbf{V}_1^{\top} &
        \cdots &
        \mathbf{V}_P^{\top}
    \end{bmatrix}^{\top}
    \in\mathbb{R}^{N\times d_t}.
\]
Identifying the tensor and flattened representations of the \(PM\) tokens, the
nonlocal component of Physics Transformer can be written compactly as
\[
    \boxed{
    \mathcal{K}_{\theta}(\mathbf{V})
    =
    \boldsymbol{\Psi}\,
    \mathcal{H}_{\theta}
    \left(
        \boldsymbol{\Phi}^{\top}\mathbf{V}
    \right)
    \mathbf{W}_o .
    }
\]
Thus, Physics Transformer realizes the sequence
\[
    N
    \xrightarrow{\ \boldsymbol{\Phi}^{\top}\ }
    PM
    \xrightarrow{\ \mathcal{H}_{\theta}\ }
    PM
    \xrightarrow{\ \boldsymbol{\Psi}\ }
    N.
\]
It therefore performs global physical interaction without explicitly
parameterizing or evaluating a dense \(N\times N\) kernel.

Conditioned on the input-dependent bases and attention weights in one forward
pass, the attention component induces an effective token-position mixing
matrix
\[
    \boldsymbol{\Gamma}_{\theta}(\mathbf{T})
    \in\mathbb{R}^{PM\times PM}
\]
for each attention head. Suppressing channel projections, its induced
point-wise mixing can be written as
\[
    \overline{\mathbf{F}}
    \approx
    \underbrace{
        \boldsymbol{\Psi}
        \boldsymbol{\Gamma}_{\theta}(\mathbf{T})
        \boldsymbol{\Phi}^{\top}
    }_{\text{adaptive projected global operator}}
    \mathbf{V}.
\]
This expression is an interpretation of the conditional linear mixing within
one forward pass; the complete model remains nonlinear because
\(\boldsymbol{\Phi}\), \(\boldsymbol{\Psi}\), and
\(\boldsymbol{\Gamma}_{\theta}\) all depend on the input features.

\subsection{Patch-Wise Basis Capacity}
\label{sec:basis_capacity}

We next formalize why Physics Transformer has access to as many as \(PM\) basis
functions. Write \(\mathbf{x}_{p,j}\) for the \(j\)-th point in patch
\(\mathcal{P}_p\). For every patch \(p\) and token index \(m\), define the
tokenization basis function
\[
    \varphi_{p,m}(\mathbf{x}_i)
    =
    \begin{cases}
        (\boldsymbol{\Phi}_p)_{j,m},
        &
        \mathbf{x}_i=\mathbf{x}_{p,j}\in\mathcal{P}_p,
        \\[2mm]
        0,
        &
        \mathbf{x}_i\notin\mathcal{P}_p,
    \end{cases}
\]
and the de-tokenization basis function
\[
    \psi_{p,m}(\mathbf{x}_i)
    =
    \begin{cases}
        (\boldsymbol{\Psi}_p)_{j,m},
        &
        \mathbf{x}_i=\mathbf{x}_{p,j}\in\mathcal{P}_p,
        \\[2mm]
        0,
        &
        \mathbf{x}_i\notin\mathcal{P}_p.
    \end{cases}
\]
The corresponding input-adaptive test and trial spaces are
\[
    \mathcal{Q}_{h}
    =
    \operatorname{span}
    \left\{
        \varphi_{p,m}
        :
        1\leq p\leq P,\,
        1\leq m\leq M
    \right\},
\]
and
\[
    \mathcal{V}_{h}
    =
    \operatorname{span}
    \left\{
        \psi_{p,m}
        :
        1\leq p\leq P,\,
        1\leq m\leq M
    \right\}.
\]

For \(p\neq q\), basis functions from patches
\(\mathcal{P}_p\) and \(\mathcal{P}_q\) have disjoint supports. Consequently,
they are orthogonal under any diagonal discrete inner product:
\[
    \left\langle
        \varphi_{p,m},
        \varphi_{q,n}
    \right\rangle_h
    =
    0,
    \qquad
    \left\langle
        \psi_{p,m},
        \psi_{q,n}
    \right\rangle_h
    =
    0.
\]
The dimensions of the two spaces are therefore
\[
    \dim\mathcal{Q}_{h}
    =
    \sum_{p=1}^{P}
    \operatorname{rank}(\boldsymbol{\Phi}_p)
    \leq PM,
\]
and
\[
    \dim\mathcal{V}_{h}
    =
    \sum_{p=1}^{P}
    \operatorname{rank}(\boldsymbol{\Psi}_p)
    \leq PM.
\]
If \(K_p\geq M\) and the \(M\) learned assignments in every patch are linearly
independent, then
\[
    \dim\mathcal{Q}_{h}
    =
    \dim\mathcal{V}_{h}
    =
    PM.
\]
Thus, even though only \(M\) physics-token types are used within each patch,
their spatially localized realizations provide up to \(PM\) basis functions
over the full domain. In contrast, a global tokenization mechanism with \(M\)
globally supported slices has rank at most \(M\).

This distinction enables the same physics-token index to capture different
patch-conditioned structures while remaining globally coupled through
\(\mathcal{H}_{\theta}\). Under this interpretation,
\(\boldsymbol{\Phi}\) defines an adaptive test space,
\(\boldsymbol{\Psi}\) defines an adaptive trial space, and
\(\mathcal{H}_{\theta}\) defines a nonlinear finite-dimensional operator
between their coefficients. Hence, Physics Transformer can be viewed as a
patch-wise Petrov--Galerkin-style neural operator:
\[
    \mathbf{V}
    \xrightarrow{\ \boldsymbol{\Phi}^{\top}\ }
    \text{test coefficients}
    \xrightarrow{\ \mathcal{H}_{\theta}\ }
    \text{updated token coefficients}
    \xrightarrow{\ \boldsymbol{\Psi}\ }
    \text{trial-space reconstruction}.
\]

\section{Experiments}

We comprehensively evaluate Physics Transformer on six representative physical field prediction benchmarks spanning 2D regular grids, irregular meshes, and 3D point clouds. These benchmarks are selected to test whether SGNO can serve as an effective general-purpose neural solver across different physical regimes, discretization structures, and PDE complexities.

\subsection{Benchmarks}

\begin{figure}[!h]
    \centering
    \includegraphics[width=1\linewidth]{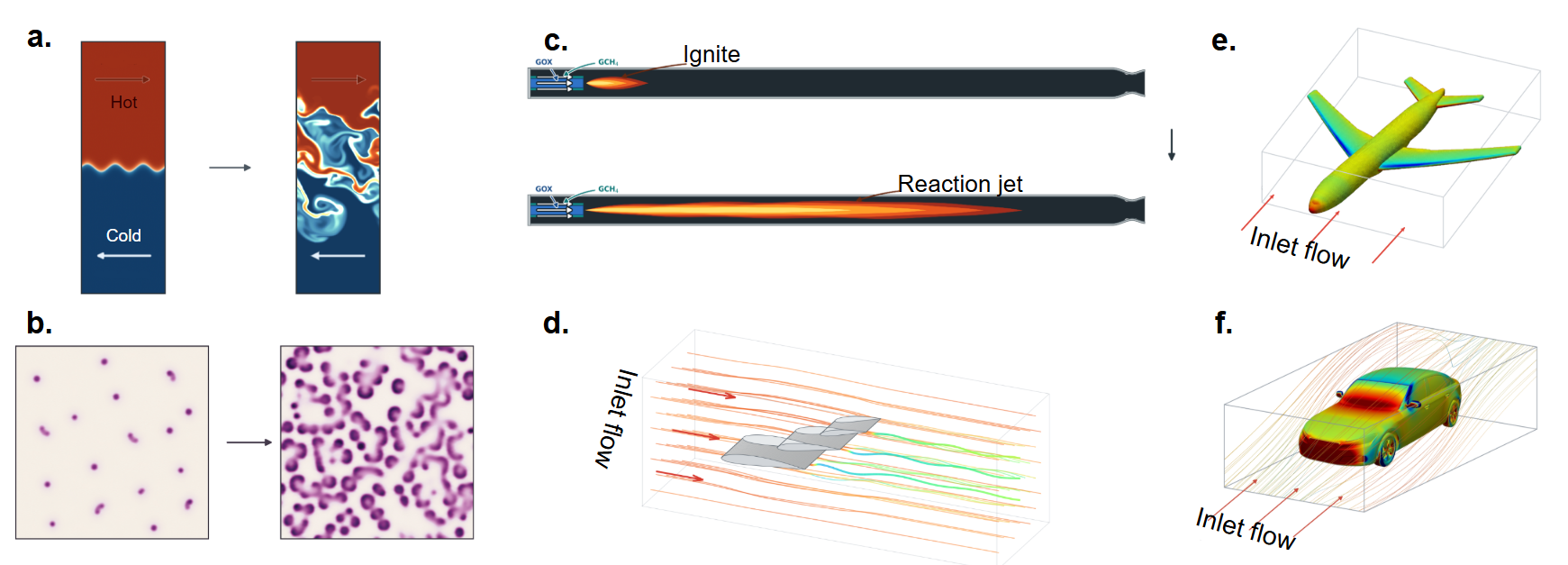}
    \caption{We test Physics Transformer on six challenging tasks covering different discretizations with complex dynamics.}
    \label{fig:task}
\end{figure}

\paragraph{2D regular-grid dynamics.}
For regular-grid evaluation, we use two datasets from The Well~\citep{ohana2024well}: \path{turbulent_radiative_layer_2D} (TRL2D) and \path{gray_scott_reaction_diffusion} (GSRD). The former describes a turbulent radiative mixing layer where cold dense gas and hot dilute gas move relative to each other, featuring rich vortical structures, multiscale temporal evolution, and localized high-frequency patterns. The Gray--Scott reaction--diffusion dataset consists of two interacting chemical species whose concentrations evolve over space and time. Depending on the feed and kill rates, the system exhibits diverse pattern-formation behavior. These benchmarks emphasizes both local dependencies and long-range coupling with fine-grained local structures, making them challenging.

\paragraph{Irregular-domain multiphysics simulations.}
For irregular-domain evaluation, we use the rocket-engine dataset from REALM~\citep{mao2025realm}. This case represents a propulsion-related reactive-flow scenario, where strong coupling, stiffness, heterogeneous scales, and mesh irregularity make accurate rollout prediction challenging. It tests whether models can operate on irregular meshes while maintaining physically meaningful predictions in regimes beyond canonical PDE settings.

We further evaluate on a geometry-grounded airflow dynamics benchmark, which we refer to as the \emph{F1-Airfoil Dynamics} benchmark. This benchmark is based on 3D airfoil geometries inspired by Formula-1 front wings and was released through the GRaM Competition at ICLR 2026~\citep{gram2026competition}. Given the underlying geometry and past velocity fields, the task is to predict future airflow dynamics. This benchmark stresses the model's ability to reason over geometry-conditioned, time-dependent flow fields on complex 3D structures.

\paragraph{Large-scale industrial CFD.}
To further examine scalability on industrial-level 3D simulations, we evaluate SGNO on two aerodynamic CFD benchmarks. First, we use the NASA Common Research Model (NASA-CRM) dataset, which contains 3D RANS simulations over a full aircraft configuration under varying flight conditions and control-surface deflections~\citep{catalani2025scalable}. This benchmark requires the model to predict aerodynamic surface fields on a complex aircraft geometry and to generalize across operating conditions. 

Second, we use DrivAerML~\citep{ashton2024drivaerml}, an industrial-scale high-fidelity automotive CFD dataset based on 500 parametrically morphed DrivAer vehicle geometries. The dataset provides rich aerodynamic fields generated by scale-resolving CFD workflows representative of industrial practice. DrivAerML poses a strong scalability challenge due to its high-resolution meshes and geometry-dependent aerodynamic structures.

\begin{table}[t]
\centering
\caption{Summary of the evaluation benchmarks. ``Rollout'' indicates whether
the task is evaluated autoregressively.}

\label{tab:benchmarks}
\tableformat
\begin{adjustbox}{max width=\linewidth}
\begin{tabular}{llcc}
\toprule
Benchmark & Discretization & Rollout & Resolution / points \\
\midrule
Turbulence 
& 2D regular grid 
& Yes 
& \(384\times128\) \\

Gray--Scott Reaction--Diffusion 
& 2D regular grid 
& Yes 
& \(128\times128\) \\

REALM Rocket Combustor 
& 2D irregular mesh 
& Yes 
& $\sim$ 250K \\

F1-Airfoil Dynamics 
& 3D point cloud 
& No
& $\sim$ 100K \\

NASA CRM 
& 3D surface mesh 
& No 
& $\sim$ 400K \\

\multirow{2}{*}{DrivAerML} 
& 3D surface mesh
& \multirow{2}{*}{No}
& $\sim$ 8M \\
& 3D volume mesh 
&
& $\sim$ 160M \\
\bottomrule
\end{tabular}
\end{adjustbox}
\end{table}

\paragraph{Baselines.}
We compare Physics Transformer with representative neural solvers from three families. 
First, we include Transolver-style neural operators, including 
Transolver~\citep{wu2024transolver}, Transolver-3~\citep{wu2026transolver3}, 
LinearNO~\citep{hu2025linearno}, and GeoTransolver~\citep{adams2025geotransolver}. 
Second, we compare with latent-space operator learners, including 
UPT~\citep{alkin2024universal}, AB-UPT~\citep{alkin2025abupt}, and GAOT~\citep{wen2025geometry}. 
Third, we include Fourier-based neural operators, including 
FNO~\citep{li2021fourier}, FFNO~\citep{tran2023factorized}, and 
GINO~\citep{li2023geometry}, to provide comparisons with classical operator-learning backbones. 
All methods are trained with the same data splits and evaluated using identical metrics. 
We use official implementations whenever available and amounts of parameters are controlled comparable. 
\subsection{Main Results}
\label{sec:main_results}

We report the main quantitative results in Tables~\ref{tab:main_results_rollout} 
and~\ref{tab:airfoil_crm_drivaerml_results}. Overall, Physics Transformer achieves consistently 
strong performance across regular-grid dynamics, irregular-domain simulations, 
and large-scale 3D aerodynamic tasks. These results demonstrate that 
Physics Transformer is not specialized to a genre of single tasks, 
but can serve as a general Transformer-based neural solver for physical field prediction.

\paragraph{Regular-grid and irregular-domain rollout prediction.}
On the Turbulent Radiative Layer, Gray--Scott reaction--diffusion, and REALM rocket-engine 
benchmarks, Physics Transformer achieves best one-step prediction accuracy and, more importantly, 
maintains stable performance under 20-step autoregressive rollout. This is crucial for 
time-dependent PDE surrogate modeling, where small one-step errors can rapidly accumulate 
and lead to physically implausible long-horizon predictions.

In contrast, Transolver and LinearNO obtain competitive one-step errors on several tasks, 
but their rollout performance degrades more rapidly. We attribute this behavior to the premature global aggregation used by transolver as a global-token neural solver. Although global 
physical tokens provide an efficient way to summarize the entire field, they may suppress 
localized fine-grained structures before temporal propagation. Once these local details 
are lost, subsequent autoregressive predictions rely increasingly on biased global 
representations, causing the rollout error to be amplified over time. By contrast, 
Physics Transformer implement local physics tokenization, preserving local 
fine-grained structures while still enabling long-range physical interaction, leading to 
more stable rollout behavior.

FFNO also provides a strong comparison on structured 2D grids. On the Turbulent Radiative 
Layer benchmark, Physics Transformer outperforms FFNO, suggesting that localized physical 
tokenization is beneficial for capturing vortical structures and multiscale temporal 
dynamics. On the Gray--Scott reaction--diffusion dataset, FFNO achieves slightly better 
performance, which is consistent with the strength of Fourier-based operators on regular grids. However, FFNO degrades substantially on the 
REALM rocket-engine task, highlighting the limitation of Fourier-based architectures when 
handling irregular meshes and complex geometries.

\begin{table}[t]
\centering
\caption{One-step and autoregressive-rollout results on time-dependent
benchmarks. Rollout error is averaged over the predicted trajectory. Lower is
better.}
\label{tab:main_results_rollout}
\tableformat
\begin{adjustbox}{max width=\linewidth}
\begin{tabular}{l|cc|cc|cc}
\toprule
\textsc{Models}
& \multicolumn{2}{c|}{\textsc{Turbulence}}
& \multicolumn{2}{c|}{\textsc{Gray--Scott}}
& \multicolumn{2}{c}{\textsc{REALM Rocket Engine}} \\
\cmidrule(lr){2-3}
\cmidrule(lr){4-5}
\cmidrule(lr){6-7}
& One-step 
& Rollout 
& One-step 
& Rollout 
& One-step 
& Rollout \\
\midrule
\textsc{FNO}
& 0.29 & 0.61
& 0.14 & 0.67
& -- & -- \\
\textsc{FFNO}
& 0.16 & 0.53
& 0.13 & 0.47
& 2.08 & 3.75 \\

\textsc{UPT}
& 0.59 & 0.81
& 1.11 & 2.05
& 0.69 & 0.92 \\

\textsc{GAOT}
& 0.23 & 0.58
& 0.28 & 0.90
& 2.37 & 2.59 \\

\textsc{Transolver}
& 0.29 & 1.07
& 0.32  & 2.09
& 0.86 & 1.37 \\

\textsc{LinearNO}
& 0.74 & 1.08
& 0.53 & 2.64
& 0.61 & 0.87 \\
\midrule
\textbf{\textsc{Physics Transformer}}
& \textbf{0.15} & \textbf{0.42}
& \textbf{0.10} & \textbf{0.49}
& \textbf{0.50} & \textbf{0.71} \\
\bottomrule
\end{tabular}
\end{adjustbox}
\end{table}

\paragraph{Geometry-dependent and large-scale CFD prediction.}
On the F1-Airfoil Dynamics, NASA-CRM, and DrivAerML benchmarks, Physics Transformer again 
achieves the best overall performance.In these tasks the physical fields are conditioned on complex 3D geometries, and accurate prediction requires the model to capture geometry-induced local flow 
structures as well as global aerodynamic interactions. The consistent improvement of 
Physics Transformer on these benchmarks indicates that its patch-wise physical tokenization is 
effective not only for temporal PDE dynamics, but also for geometry-conditioned field prediction.

GeoTransolver also performs competitively on these geometry-related tasks, benefiting from its explicit modeling of geometric information. However, its design is primarily tailored 
to geometry-aware prediction. In contrast, Physics Transformer does not rely on a geometry-specific architecture, yet still achieves better performance on geometry-dependent tasks. This suggests that Physics Transformer provides a more general mechanism for 
accurate physical fields. Transolver-3 further improves over previous 
Transolver variants by scaling to larger geometries, but its global aggregation backbone 
still limits its ability to handle local interactions, leaving room for 
Physics Transformer to achieve stronger accuracy.

\begin{table}[t]
\centering
\caption{Comparison on F1-Airfoil, NASA-CRM, and DrivAerML benchmarks.
Lower error is better. The best results are highlighted in bold and the second-best results are underlined.}
\label{tab:airfoil_crm_drivaerml_results}
\tableformat
\begin{adjustbox}{max width=\linewidth}
\begin{tabular}{l|c|cc|cccc}
\toprule
\textsc{Models}
& \multicolumn{1}{c|}{\textsc{F1-Airfoil}}
& \multicolumn{2}{c|}{\textsc{NASA-CRM}}
& \multicolumn{4}{c}{\textsc{DrivAerML}} \\
\cmidrule(lr){2-2}
\cmidrule(lr){3-4}
\cmidrule(lr){5-8}
& $u$
& $p_s$ & $C_f$
& $p_s$ & $u$ & $\tau$ & $p_v$ \\
\midrule
\textsc{GINO}
& 17.71
& 12.39 & 11.51
& 13.03 & 40.58 & 21.71 & 44.90 \\

\textsc{GAOT}
& 28.46
& 30.38 & 59.79
& 34.00 & 57.18 & 61.00 & 56.90 \\

\textsc{UPT}
& 13.27
& 12.78 & 23.78
& 7.44 & 8.74 & 12.93 & 10.05 \\

\textsc{AB-UPT}
& 6.38
& 9.77 & 6.43
& 3.82 & 5.93 & 7.29 & 6.08 \\

\textsc{Transolver}
& 5.50
& 9.61 & 7.04
& 4.81 & 6.78 & 8.95 & 7.74 \\

\textsc{Transolver-3}
& --
& 8.71 & 5.85
& 3.71 & \underline{4.14} & 5.85 & 5.72 \\

\textsc{GeoTransolver}
& \underline{4.45}
& \underline{6.89}
& \underline{4.85}
& \underline{3.64}
& 4.67
& \underline{5.71}
& \underline{4.05} \\
\midrule
\textbf{\textsc{Physics Transformer}}
& \textbf{3.98}
& \textbf{5.82}
& \textbf{3.31}
& \textbf{3.35}
& \textbf{3.40}
& \textbf{5.24}
& \textbf{3.08} \\
\bottomrule
\end{tabular}
\end{adjustbox}
\end{table}

\paragraph{Integrated aerodynamic quantities on DrivAerML.}
In industrial CFD applications, pointwise field accuracy is important, but integrated 
aerodynamic quantities are often even more directly related to engineering decisions such as the drag coefficient $C_d$ and lift coefficient $C_l$. These quantities are obtained by integrating predicted 
surface fields over the vehicle geometry, and therefore evaluate whether a surrogate model 
can preserve physically meaningful global aerodynamic responses, rather than merely 
matching local pointwise values.

To assess this capability, we compute the coefficient of determination $R^2$ for $C_d$ 
and $C_l$ on DrivAerML, comparing Physics Transformer with GeoTransolver and Transolver-3. 
As shown in Table~\ref{tab:drivaerml_integral}, 
Physics Transformer achieves $R^2$ scores above $0.99$ for both $C_d$ and $C_l$, indicating that 
it can accurately recover industrially relevant aerodynamic quantities..

\begin{table}[!h]
\centering
\caption{$R^2$ prediction accuracy of integrated aerodynamic quantities on DrivAerML. 
Higher is better.}
\label{tab:drivaerml_integral}
\tableformat
\begin{tabular}{lcc}
\toprule
Model & $C_d$ & $C_l$ \\
\midrule
Transolver-3 & 0.972 & 0.985 \\
GeoTransolver &  $\mathbf{0.996}$  & 0.991 \\
Physics Transformer & $\mathbf{0.996}$ & $\mathbf{0.993}$ \\
\bottomrule
\end{tabular}
\end{table}

\paragraph{Summary.}
Across representative physical prediction tasks, Physics Transformer consistently demonstrates 
strong accuracy, stable rollout behavior, and favorable scalability. Its success on both time-dependent PDEs and industrial-scale aerodynamic simulations shows that localized physics tokenization and efficient global interaction are essential for 
building a general-purpose Transformer architecture for physical field prediction.

\subsection{Effect of Point Ordering for Patchification}
\label{sec:ordering_ablation}

For irregular meshes and point-cloud inputs, Physics Transformer first partitions points into local patches and then performs patch-wise physical tokenization. 
The quality and cost of this patchification step depend on the point ordering or grouping strategy. 
We compare three representative choices: Morton ordering, Hilbert ordering, and BallTree-based partitioning.

Morton ordering is computationally the cheapest among the three, as it maps multidimensional coordinates to one-dimensional indices through bit interleaving. 
However, although Morton ordering is locality-preserving in a coarse sense, it may introduce discontinuities in the one-dimensional traversal, causing spatially distant points to be assigned to the same patch. 
Such discontinuities can mix physically unrelated local regions and make the resulting patch tokens semantically ambiguous. 
Hilbert ordering incurs slightly higher preprocessing cost, but it provides better spatial continuity along the traversal curve and therefore produces more coherent local patches. 
BallTree-based partitioning constructs compact spatial groups through a tree-based nearest-neighbor structure. 
It usually gives the most geometrically compact patches, but its preprocessing cost is significantly higher.

To quantify this trade-off, we evaluate different patchification strategies on DrivAerML. 
For each method, we report the sorting or grouping time for 100K points and the corresponding prediction performance. 
As shown in Table~\ref{tab:ordering_ablation}, Morton ordering leads to a clear performance drop, indicating that discontinuous patches can harm local physical tokenization. 
BallTree partitioning provides only marginal improvement over Hilbert ordering, despite its much higher preprocessing cost. 
This suggests that Physics Transformer does not rely on overly regular or perfectly compact patch partitions; instead, its physics tokenization learns robust local physical representations as long as the patches preserve reasonable spatial coherence. 
Considering both efficiency and accuracy, we use Hilbert ordering as the default patchification strategy in all experiments.

\begin{table}[h]
\centering
\caption{Effect of point-ordering strategies on DrivAerML. We report
preprocessing latency for \(100\)K points and downstream relative \(L_2\)
errors. Lower is better.}
\label{tab:ordering_ablation}
\tableformat
\begin{tabular}{lccc}
\toprule
Grouping Strategy 
& Latency 
& $p_s$
& $\tau$
 \\
\midrule
Morton ordering 
& 12 ms
& 3.98 
& 5.96 \\

Hilbert ordering 
& 50 ms
& 3.35
& 5.24 \\

BallTree partitioning 
& 479 ms
& 3.33 
& 5.25 \\
\bottomrule
\end{tabular}
\end{table}

\subsection{Effect of the Number of Patches}
\label{sec:num_patches_ablation}

The number of patches controls the granularity of local physical tokenization in Physics Transformer. 
A smaller number of patches leads to larger local regions, which improves computational efficiency but may mix heterogeneous local structures into the same patch. 
A larger number of patches provides finer local decomposition, but also reduces the number of points within each patch and increases the cost of global token interaction. 
Therefore, the patch number introduces a trade-off between local detail preservation, projection stability, and computational efficiency.

We study this effect on two representative benchmarks: DrivAerML and Turbulent Radiative Layer. 
DrivAerML contains relatively smooth vehicle surface fields with strong geometry dependence, while TRL contains rich vortical structures and localized multiscale dynamics. 
These two tasks therefore have different requirements on the granularity of local physical representation. 
We vary the number of patches from $128$ to $4096$ and report the corresponding prediction errors in Table~\ref{tab:num_patches_ablation}.

\begin{table}[h]
\centering
\caption{Effect of the number of patches on DrivAerML and TRL. 
Lower error is better. The best result for each setting is highlighted in bold.}
\label{tab:num_patches_ablation}
\tableformat
\begin{adjustbox}{max width=\linewidth}
\begin{tabular}{lcccccc}
\toprule
Benchmark 
& 128 
& 256 
& 512 
& 1024 
& 2048 
& 4096 \\
\midrule
DrivAerML -- $p_s$
& 4.02
& 3.64
& 3.45
& 3.35
& 3.32 
& \textbf{3.30} \\

DrivAerML -- $\tau$
& 6.13 
& 5.58
& 5.31
& 5.24
& 5.22 
& \textbf{5.19} \\

TRL -- One-step
& 0.22 
& 0.18 
& 0.16 
& \textbf{0.15} 
& \textbf{0.15} 
& 0.17 \\

TRL -- Rollout
& 0.55
& 0.47
& 0.43
& 0.42
& 0.43 
& 0.46 \\
\bottomrule
\end{tabular}
\end{adjustbox}
\end{table}

The results are consistent with our expectation. 
For both benchmarks, increasing the number of patches initially improves performance, indicating that finer local decomposition helps the model learn more accurate local physical representations. 
On TRL, the improvement is substantial when increasing the number of patches from $128$ to $1024$, reflecting the importance of fine-grained locality for turbulent and multiscale dynamics. 
However, the gain becomes marginal at $2048$ patches, and performance even slightly degrades at $4096$ patches. 
We conjecture that overly fine partitioning reduces the number of points available within each patch, making the learned local basis functions under-sampled and the patch-wise projection less stable. 

On DrivAerML, the performance saturates earlier: increasing the patch number beyond $512$ brings only minor improvement. 
This is consistent with the smoother nature of vehicle surface fields, where a moderate number of patches is already sufficient to capture geometry-conditioned local variations. 
Overall, the benefit of increasing the patch number exhibits a task-dependent saturation behavior.

Importantly, the global token attention in Physics Transformer scales quadratically with the number of patches. Therefore, using an excessively large number of patches can substantially increase computation without bringing meaningful accuracy gains. 
Based on the trade-off observed across DrivAerML and TRL, we use $1024$ patches as the default setting in our main experiments, which provides a robust balance between accuracy and efficiency.
\subsection{Effect of the Number of Basis Functions}
\label{sec:num_basis_ablation}

We further study the effect of the number of basis functions, i.e., the number of learned physics tokens per patch. 
Unlike the number of patches, which changes the spatial granularity of the domain decomposition, the number of basis functions controls the capacity of the local function-projection space within each patch. 
A larger number of basis functions allows the model to represent richer local physical variations, but it also increases the cost of tokenization, global token interaction, and de-tokenization.

We vary the number of basis functions from $16$ to $128$ and evaluate the model on DrivAerML and Turbulent Radiative Layer. 
The results are reported in Table~\ref{tab:num_basis_ablation}.

\begin{table}[h]
\centering
\caption{Effect of the number of basis functions on DrivAerML and TRL. 
The number of basis functions refers to the number of learned physics tokens per patch. 
Lower error is better.}
\label{tab:num_basis_ablation}
\tableformat
\begin{tabular}{lcccc}
\toprule
Benchmark 
& 16 
& 32 
& 64 
& 128 \\
\midrule
DrivAerML -- $p_s$
& 3.43 
& 3.39 
& 3.35 
& 3.34 \\

DrivAerML -- $\tau$
& 5.31
& 5.27
& 5.24 
& 5.22 \\

TRL -- One-step
& 0.155 
& 0.152
& 0.151 
& 0.151 \\

TRL -- Rollout
& 0.443 
& 0.421 
& 0.419 
& 0.417 \\
\bottomrule
\end{tabular}
\end{table}

As shown in Table~\ref{tab:num_basis_ablation}, increasing the number of basis functions consistently improves the model performance, which indicates that a richer local projection basis helps encode more detailed physical structures within each patch. 
However, the improvement from more basis functions is relatively small. 
Since the token-mixing complexity grows with the number of basis functions,
we set the number of basis functions to $32$ or $64$ in our main experiments, depending on the complexity of the physical problem.
\subsection{Effect of Mesh Scale}
\label{sec:point_scale}

\begin{figure}[!t]
    \centering
    \includegraphics[width=1\linewidth]{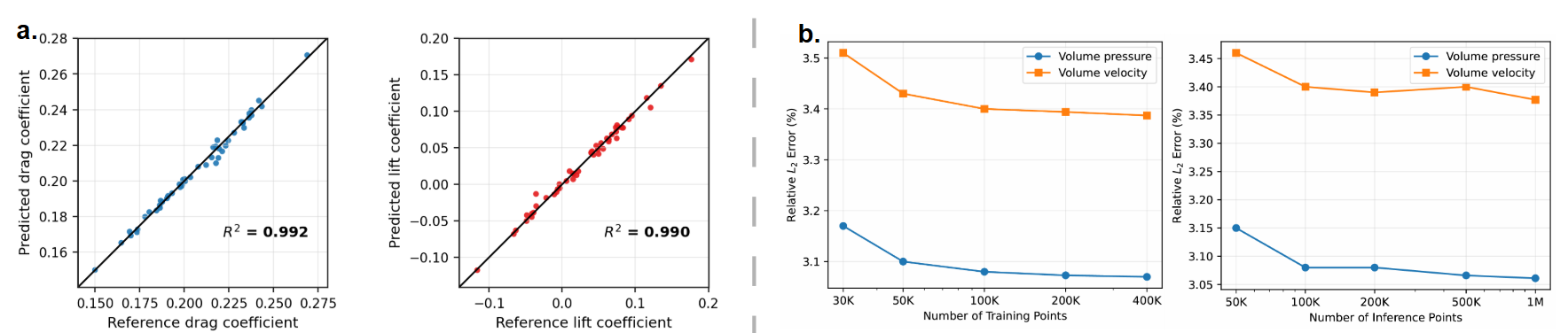}
    \caption{In part a, we show the satisfying predication accuracy of integrated quantities of Physics Transformer on DrivAerML. In part b, we compare the performance of Physics Transformer under different mesh density during traning and inference. The model used for inference is trained with 100k mesh points. It could be seen the performance saturate when using more than 100k points. }
    \label{fig:line}
\end{figure}

DrivAerML contains million-scale surface meshes and even larger volumetric CFD grids, making it impractical to train neural solvers on all points. 
Moreover, compared with mesh density, the vehicle geometry and its surface aerodynamic fields are relatively smooth, suggesting that a moderate number of sampled points may already capture the dominant geometry-dependent physical patterns. 
This is consistent with the goal of neural solvers: rather than relying on extremely dense numerical discretization for convergence as in classical CFD solvers, they learn a fast mapping from input geometry to output physical fields.
We first vary the number of training points sampled from each full point cloud, using $30$K, $50$K, $100$K, $200$K, and $400$K points. 
In this ablation, the relative $L_2$ error decreases as the training point
count increases, but the improvement gradually saturates after $50$K points.
Using more points substantially increases memory and computation, while bringing only marginal accuracy gains. 
Therefore, we use $100$K randomly sampled points for training in our experiments.

We further evaluate the effect of inference point count using the model trained with $100$K points. 
We increase the input point scale from $50$K to $1$M points. Increasing the inference scale from $100$K to $1$M improves the relative $L_2$ error by less than $1\%$. 
Both ablations indicate that Physics Transformer can already capture the vehicle geometry and geometry-dependent aerodynamic fields with a moderate number of points, while full-resolution predictions can be efficiently obtained by chunk-wise inference.

\begin{figure}[!t]
    \centering
    \includegraphics[width=1\linewidth]{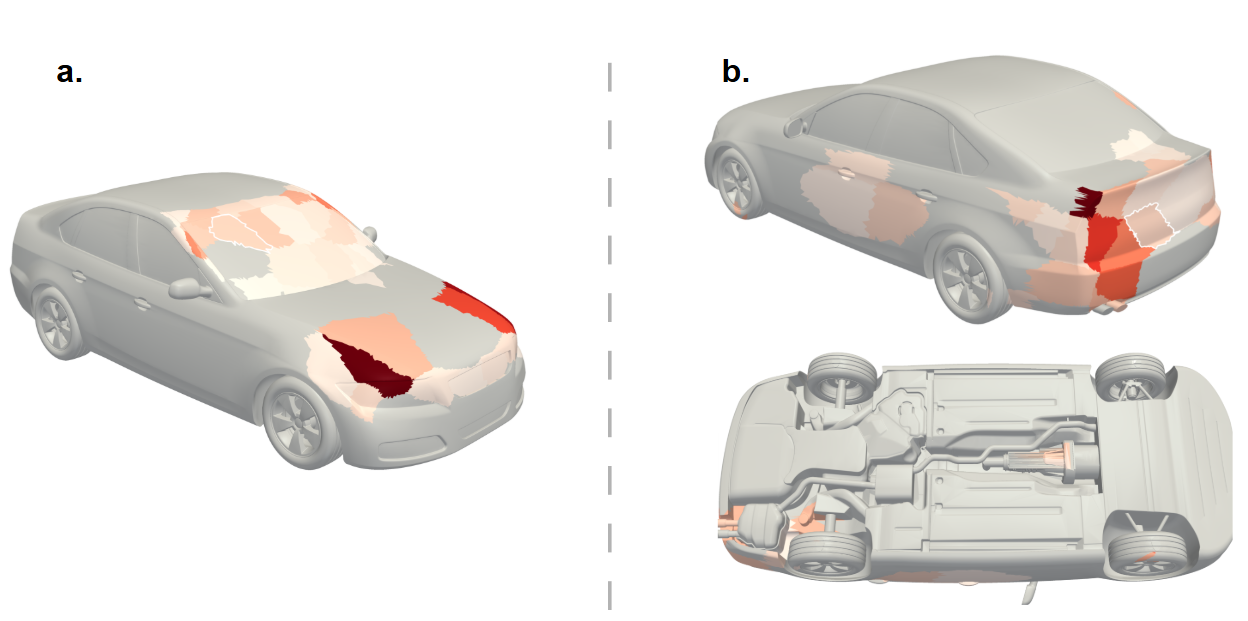}
    \caption{We illustrate two representative attention scores from the last layer of Physics Transformer among 1024 patches. The query token is circled by white line. Deeper red color express higher attention score. We draw only the top 5\% patches. It could be seen that Physics Transformer could capture meaningful local interactions and long-range dependency. }
    \label{fig:att}
\end{figure}

\subsection{Comparison with Point-Cloud Patchification Methods}
\label{app:comparison_point_patchification}

Point Transformer V3 (PTv3)~\citep{wu2024pointtransformerv3} and
SpiderSolver~\citep{qi2025spidersolver} both organize computation through
spatial partitioning of irregular point sets. Although neither provides a
unified neural solver across all the discretization types considered in our
work, they can be adapted to the three-dimensional point-cloud representations
used by DrivAerML. We therefore compare them with Physics Transformer on this
industrial-scale automotive CFD benchmark. For PTv3, we adopt its largest
model configuration and replace the original perception head with a
point-wise field-regression head. We also scale the width and depth of
SpiderSolver to approximately \(30\) million trainable parameters, ensuring
that its performance is not restricted by insufficient model capacity. All use 100k points to train. The
results are reported in Table~\ref{tab:point_patchification_comparison}.

\begin{table}[h]
    \centering
    \caption{
        Comparison with point-cloud patchification methods on DrivAerML.
        Lower error is better.
    }
    \label{tab:point_patchification_comparison}
    \tableformat
    \begin{adjustbox}{max width=\linewidth}
    \begin{tabular}{lcccc}
        \toprule
        Method
        & $p_s$
        & $u$
        & $\tau$
        & $p_v$ \\
        \midrule
        Point Transformer V3~\citep{wu2024pointtransformerv3}
        & 15.53
        & 17.71
        & 19.34
        & 20.69 \\

        SpiderSolver~\citep{qi2025spidersolver}
        & 7.48
        & 14.28
        & 10.54
        & 8.13 \\

        Physics Transformer (Ours)
        & 3.35
        & 3.40
        & 5.24
        & 3.08 \\
        \bottomrule
    \end{tabular}
    \end{adjustbox}
\end{table}

\paragraph{Discussion.}
The main difference among these methods lies not in how the spatial patches
are constructed, but in how the physical field within each region is
represented by tokens. SpiderSolver employs a boundary-guided spiderweb
partition that is well suited to complex CFD geometries. However, it constructs
each spiderweb token by average-pooling the point features within the
corresponding subregion. Consequently, each subregion is represented by a
single coarse summary that primarily captures its mean feature statistics.
Heterogeneous physical states and fine-scale variations within the same
subregion may therefore be discarded before global token interaction, and
cannot be recovered by subsequent attention. Its fine-grained branch enhances
interactions between boundary points and nearby interior points, but does not
provide multiple independent representation modes for general field variations
within each subregion.

PTv3 exhibits a different limitation. It retains serialized point or voxel
features as tokens and uses the resulting patches primarily as local attention
windows. Its hierarchical representations are formed through geometric
downsampling and local feature aggregation originally designed for semantic
3D perception. Thus, PTv3 does not explicitly project a continuous physical
field onto compact local functional modes. Long-range physical coupling must
instead emerge through repeated local attention and hierarchical propagation,
which is less effective for resolving globally coupled and multiscale PDE
fields.

In contrast, Physics Transformer learns input-adaptive basis functions within
each patch and constructs its physics tokens through the functional projection.
Rather than compressing a region into a single average representation, these
projection coefficients can preserve multiple coexisting physical states
and spatial variation modes within the same patch. The resulting patch-wise
physics tokens are then explicitly coupled across the full domain through
factorized global attention. Importantly, this functional tokenization does not depend on CFD-specific boundary assumptions and can be applied without
architectural modifications to regular grids, irregular meshes, and point
clouds. The performance advantage of Physics Transformer therefore primarily stems
from its field-adaptive token representation, rather than from the particular
spatial partitioning strategy.

\bibliographystyle{plainnat}
\bibliography{references}

\newpage
\appendix

\section{Full Mesh Inference}
\subsection{Arbitrary-Point Querying}
\label{app:arbitrary_query}

\noindent\textbf{Notation.}
\(\mathcal{S}=\{(\mathbf{x}_i,\mathbf{a}_i)\}_{i=1}^{N_s}\) denotes the
support set used to encode the input function, where \(\mathbf{X}_s\) and
\(\mathbf{A}_s\) collect its coordinates and input features.
\(\mathcal{Q}=\{(\mathbf{x}_q,\mathbf{a}_q)\}_{q=1}^{N_q}\) denotes the query
set. The support locations may be a subset
of or different from the query locations, and typically \(N_s\ll N_q\).
\(L\), \(P\), and \(M\) denote the numbers of network layers, spatial patches,
and physics tokens per patch, respectively. At layer \(\ell\),
\(\mathbf{F}_s^\ell\) denotes the support-point features,
\(\mathbf{T}^\ell\in\mathbb{R}^{P\times M\times d_t}\) denotes the updated
physics tokens after factorized global attention, \(p_i^\ell\) denotes the
patch index of support point \(\mathbf{x}_i\), and \(\sigma_\ell\) denotes the
axis permutation used for Trans-Hilbert patchification. The cache
\(\mathcal{C}\) stores the physics tokens and patch assignments from all
layers.

\subsection{Resolution Agnostic Property}
\label{app:arbitrary_point_querying}

We evaluate the arbitrary-point querying capability of Physics Transformer on
DrivAerML using 50 test cases. We first construct and cache the patch-wise physics tokens at all
layers using a support set of only \(100\)K mesh points. We then keep these
cached tokens fixed and query the predicted fields independently at every point
of the full-resolution mesh. Table~\ref{tab:arbitrary_point_querying} compares
prediction errors evaluated on the \(100\)K support set and on the complete mesh
through arbitrary-point querying.

\begin{table}[h]
    \centering
    \caption{
        Arbitrary-point querying results on DrivAerML. The physics tokens are
        computed once from a \(100\)K-point support set and reused to query the
        full-resolution mesh. Lower error is better.
    }
    \label{tab:arbitrary_point_querying}
    \tableformat
    \begin{tabular}{lcccc}
        \toprule
        \textsc{Evaluation Mode}
        & \(p_s\)
        & \(\mathbf{u}\)
        & \(\boldsymbol{\tau}\)
        & \(p_v\) \\
        \midrule
        Support-set prediction (\(100\)K)
        & 3.35& 3.40& 5.24& 3.08\\
        Cached-token querying (full mesh)
        & 3.34& 3.40& 5.25& 3.08\\
        \bottomrule
    \end{tabular}
\end{table}

Querying the complete mesh from the cached tokens introduces no measurable
accuracy degradation relative to prediction on the original support set. This
result demonstrates that Physics Transformer exhibits satisfactory
resolution-independent behavior (super resolution only): its patch-wise physics tokens form an
efficient representation of the local physical fields that can be decoded at
previously unseen coordinates without recomputing the support-set
representation. Consequently, a relatively small support set is sufficient to
encode the physical state, while predictions can subsequently be evaluated on a
substantially denser mesh or obtain planar cross-sections of the 3D domain for visualization.

\subsection{Difference from Transolver-3 Tiled Inference}
\label{app:comparison_transolver3_tiling}

Although Transolver-3 provides a point-wise decoding rule, its practical
inference pipeline still constructs the physical-state cache from the whole
high-resolution mesh and subsequently performs full-mesh decoding. Importantly,
the ability to decode an individual point does not imply that the physical
representation can be reliably constructed from a sparse support set. We test Transolver-3~\citep{wu2026transolver3} by reducing the number of
input cells used for physical-state caching from the full mesh to \(100\)K 
increases the prediction error from \(\mathrm{ 3.71\%,4.14\%,5.85\%,5.72\%}\) ($p_s, u, \tau, p_v$ respectively) to
\(\mathrm{3.85\%,4.33\%,6.16\%,6.09\%}\). This substantial degradation suggests that its global
slice aggregation is relatively sample-inefficient: each physical state pools
information over the entire domain, and sparse sampling may provide
insufficient coverage of localized geometry and fine-scale physical
structures.

In contrast, Physics Transformer constructs physics tokens independently within
spatial patches. Because each token only needs to represent the local geometry
and physical variation inside its associated patch, fine-scale information is
less likely to be diluted by aggregation over distant regions. Consequently,
Physics Transformer can construct an accurate physical representation from a
relatively small support set
\(\mathcal{S}\), cache the resulting layer-wise patch tokens, and efficiently
decode the solution on a much denser query discretization
\(\mathcal{Q}\), where \(N_s\ll N_q\). Therefore, the key distinction is not
whether a single point can be decoded from cached states, but how many support
points are required to construct a sufficiently informative tokens for decoding.

\begin{algorithm}[t]
\caption{Arbitrary-point querying with cached physics tokens}
\label{alg:arbitrary_query}
\begin{algorithmic}[1]
\Require Support set
\(\mathcal{S}=\{(\mathbf{x}_i,\mathbf{a}_i)\}_{i=1}^{N_s}\);
query set
\(\mathcal{Q}=\{(\mathbf{x}_q,\mathbf{a}_q)\}_{q=1}^{N_q}\);
number of layers \(L\)
\Ensure Query predictions
\(\widehat{\mathcal{U}}_{\mathcal{Q}}
=\{\widehat{\mathbf{u}}(\mathbf{x}_q)\}_{q=1}^{N_q}\)

\State
\(
\mathbf{F}_s^0
\gets
\phi_{\rm in}
\left(
    [\mathbf{A}_s,\gamma(\mathbf{X}_s)]
\right)
\)
\Comment{Initial support-point embeddings}

\State
\(\mathcal{C}\gets [\space]\)
\Comment{Multi-layer token cache}

\For{\(\ell=1,\ldots,L\)}
    \State
    \(
    \left(
        \mathbf{F}_s^\ell,
        \mathbf{T}^\ell,
        \{p_i^\ell\}_{i=1}^{N_s}
    \right)
    \gets
    \Call{PhysicsTransformerLayer}
    {
        \mathbf{X}_s,
        \mathbf{F}_s^{\ell-1},
        \sigma_\ell
    }
    \)
    \Comment{Patchify, tokenize, factorized-attention, and de-tokenize}

    \State
    \(
    \mathcal{C}[\ell]
    \gets
    \left(
        \mathbf{T}^\ell,
        \{p_i^\ell\}_{i=1}^{N_s}
    \right)
    \)
    \Comment{Cache tokens and patch indices at layer \(\ell\)}
\EndFor

\ForAll{\((\mathbf{x}_q,\mathbf{a}_q)\in\mathcal{Q}\)}
    \State
    \(
    \mathbf{f}_q^0
    \gets
    \phi_q
    \left(
        [\gamma(\mathbf{x}_q),\mathbf{a}_q]
    \right)
    \)
    \Comment{Initial query-point feature}

    \State
    \(
    n(q)
    \gets
    \displaystyle
    \arg\min_{1\leq i\leq N_s}
    \left\|
        \mathbf{x}_q-\mathbf{x}_i
    \right\|_2
    \)
    \Comment{Nearest support-point index}

    \For{\(\ell=1,\ldots,L\)}
        \State
        \(
        p^\ell(q)
        \gets
        p_{n(q)}^\ell
        \)
        \Comment{Query-associated patch at layer \(\ell\)}

        \State
        \(
        \mathbf{q}_q^\ell
        \gets
        \operatorname{softmax}_{M}
        \left(
            \mathbf{f}_q^{\ell-1}
            \mathbf{W}_q^\ell
        \right)
        \in\mathbb{R}^{M}
        \)
        \Comment{Layer-\(\ell\) de-tokenization weights}

        \State
        \(
        \mathbf{f}_q^\ell
        \gets
        (\mathbf{q}_q^\ell)^\top
        \mathbf{T}_{p^\ell(q)}^\ell
        \mathbf{W}_o^\ell
        \)
        \Comment{Decode from layer-\(\ell\) cached tokens}
    \EndFor

    \State
    \(
    \widehat{\mathbf{u}}(\mathbf{x}_q)
    \gets
    \phi_{\rm out}
    \left(
        \mathbf{f}_q^L
    \right)
    \)
\EndFor

\State
\Return
\(
\widehat{\mathcal{U}}_{\mathcal{Q}}
\)
\end{algorithmic}
\end{algorithm}

\section{Benchmarks}
\label{app:benchmarks}

\paragraph{Turbulent Radiative Layer 2D (TRL2D).}
TRL2D is a two-dimensional hydrodynamic dataset from The Well
benchmark~\citep{ohana2024well}. It models a cold, dense gas moving relative to
a hot, dilute gas. Their interface develops a Kelvin--Helmholtz instability,
while radiative cooling acts on gas at intermediate temperatures. The dataset
contains 90 trajectories generated from nine cooling-time settings and ten
random seeds, with each trajectory consisting of 101 time steps on a
\(384\times128\) Cartesian grid. The predicted physical fields include density,
pressure, and velocity. Given the physical fields from the previous three time
steps, the task is to predict the fields at the next time step. We adopt the
official trajectory-level split from The Well, resulting in 72 training, 9
validation, and 9 test trajectories.

\paragraph{Gray--Scott Reaction--Diffusion (GSRD).}
GSRD is another regular-grid dataset from The Well
benchmark~\citep{ohana2024well}. It describes the nonlinear reaction and
diffusion of two chemical species, whose dynamics produce spatial patterns
under different feed and removal parameters. The dataset contains 1,200
trajectories covering six parameter settings and multiple randomized initial
conditions. Each trajectory contains 1,001 time steps on a
\(128\times128\) periodic grid. Since the dynamics between consecutive time
steps evolve slowly, we temporally subsample each trajectory by a factor of 10
to obtain more informative prediction intervals. Given the concentration fields
from the previous three subsampled time steps, the task is to predict the fields
at the next subsampled time step. Following the official partition, we use 960
trajectories for training, 120 for validation, and 120 for testing.

\paragraph{REALM rocket-engine simulations.}
REALM is a collection of high-fidelity multiphysics simulations designed to
represent realistic engineering systems~\citep{mao2025realm}. We use its
rocket-engine benchmark, which models turbulent reacting flow in a rocket
combustor with shear-coaxial injection. The interaction among fuel injection,
turbulent mixing, combustion, heat release, and chamber-flow dynamics produces
strongly coupled structures over irregular discretizations. Given the physical
fields from the previous three time steps, the task is to predict the fields at
the next time step. The dataset used in our experiments contains 100 complete
trajectories. As no official partition is provided for this 100-trajectory set,
we divide it into 80 training, 10 validation, and 10 test trajectories.

\paragraph{F1-Airfoil Dynamics.}
F1-Airfoil Dynamics is the official benchmark of the GRaM Competition at
ICLR 2026, with simulated transient airflow provided by BeyondMath. It consists
of irregular point-cloud simulations around three-dimensional geometries
inspired by Formula-1 front wings. Given the geometry and five observed velocity
frames, the task is to predict the subsequent five frames jointly. Since all
future frames are produced in a single forward pass and no prediction is fed
back into the model, this benchmark does not require autoregressive rollout. We
therefore treat it as a fixed-horizon operator-learning problem rather than a
rollout-based temporal benchmark. We follow the official training, validation,
and test partition provided by the competition.

\paragraph{NASA Common Research Model (NASA-CRM).}
NASA-CRM is a transonic aerodynamic benchmark based on a full
wing--body--horizontal-tail transport-aircraft configuration
~\citep{catalani2025scalable}. It contains 149 high-fidelity RANS simulations
generated with the DLR TAU solver. Each simulation varies six parameters: Mach
number, angle of attack, inboard and outboard aileron deflections, elevator
deflection, and horizontal-tailplane deflection, while the flight altitude is
fixed at \(37{,}000\) ft. Each case contains 454,404 irregularly distributed
surface points, and the task is to predict the surface pressure coefficient over
the complete aircraft geometry. We follow the official Halton-sequence-based
partition, using 105 simulation cases for training and 44 cases for testing.

\paragraph{DrivAerML.}
DrivAerML is an industrial-scale automotive aerodynamics dataset containing
500 parametrically morphed DrivAer vehicle configurations
~\citep{ashton2024drivaerml}. The simulations are generated using
scale-resolving CFD workflows and provide high-resolution surface and volume
fields, including pressure, velocity, and wall shear stress. A typical case
contains approximately eight million surface cells and can contain up to
approximately 160 million volume cells, making DrivAerML particularly
challenging for point-wise neural solvers. Since the original dataset does not
prescribe a unique learning partition for this task, we split the 500 vehicle
configurations into 400 training, 50 validation, and 50 test cases.
\subsection{Metrics}
\label{app:metrics}

We follow the standard evaluation protocols associated with each benchmark.
Specifically, we use the variance-scaled root mean squared error (VRMSE) for
the two datasets from The Well, normalized MSE for REALM, and relative \(L_2\)
error for F1-Airfoil Dynamics, NASA-CRM, and DrivAerML. For DrivAerML, we
additionally evaluate the integrated drag and lift coefficients and the corresponding coefficient of determination \(R^2\).

\paragraph{Variance-scaled RMSE on The Well.}
Following the official evaluation protocol of The Well
~\citep{ohana2024well}, we evaluate TRL2D and GSRD using VRMSE. Let
\(\widehat{\mathbf{y}}_{c}^{(s,h)}\) and
\(\mathbf{y}_{c}^{(s,h)}\) denote the predicted and ground-truth values of
field channel \(c\) for sample \(s\) at rollout step \(h\). The VRMSE of this
field is
\[
    \operatorname{VRMSE}_{c}^{(s,h)}
    =
    \sqrt{
        \frac{
            \frac{1}{N}
            \sum_{i=1}^{N}
            \left(
                \widehat{y}_{i,c}^{(s,h)}
                -
                y_{i,c}^{(s,h)}
            \right)^2
        }{
            \operatorname{Var}_{i}
            \left[
                y_{i,c}^{(s,h)}
            \right]
            +
            \varepsilon
        }
    },
\]
where the variance is computed over the spatial domain and
\(\varepsilon=10^{-7}\) prevents numerical instability. The reported error is
averaged over all predicted physical-field channels and test samples.

The physical variables in The Well have different units and substantially
different amplitudes. Normalizing each field by its spatial standard deviation
prevents high-variance fields from dominating the aggregate error and produces
a dimensionless metric that is comparable across density, pressure, velocity,
and chemical concentration fields. VRMSE is also calibrated such that predicting
the spatial mean of the target field gives a score of approximately \(1\), while
a perfect prediction gives \(0\).

\paragraph{Normalized MSE on REALM.}
We follow the official REALM protocol~\citep{mao2025realm} and use normalized MSE
for both training and evaluation. Reactive-flow variables exhibit severe
cross-channel scale disparity: in particular, chemical-species mass fractions
can range from approximately \(10^{-12}\) to \(10^{-1}\). Directly optimizing
an error on the raw variables would therefore produce an ill-conditioned
objective dominated by variables with large numerical magnitudes.

Following REALM, the chemical-species channels are first transformed using a
channel-wise Box--Cox transformation with \(\lambda=0.1\), and all channels are
then standardized using statistics computed from the training set. Let
\(\widetilde{\mathbf{y}}_{\alpha}^{(s,h)}\) and
\(\widehat{\widetilde{\mathbf{y}}}_{\alpha}^{(s,h)}\) denote the normalized
ground truth and prediction for physical-variable group
\[
    \alpha
    \in
    \mathcal{A}
    =
    \{
        \mathrm{chem},\,T,\,\rho,\,\mathbf{u},\,p
    \}.
\]
The normalized MSE at step \(h\) is
\[
    \operatorname{MSE}_{\rm REALM}^{(s,h)}
    =
    \sum_{\alpha\in\mathcal{A}}
    \frac{1}{D_{\alpha}^{(s)}}
    \left\|
        \widehat{\widetilde{\mathbf{y}}}_{\alpha}^{(s,h)}
        -
        \widetilde{\mathbf{y}}_{\alpha}^{(s,h)}
    \right\|_{F}^{2},
\]
where \(D_{\alpha}^{(s)}\) is the number of spatial and channel elements in
group \(\alpha\). Averaging within each group prevents physical groups with
more channels from dominating the loss. We use this same group-wise MSE during
training and testing to maintain consistency between model optimization and
benchmark evaluation.

\paragraph{Autoregressive rollout evaluation.}
For TRL2D, GSRD, and REALM, each model receives the previous three frames and
predicts the next frame. During autoregressive evaluation, the prediction is
appended to the input window and the oldest frame is discarded. This procedure
is repeated for \(H\) steps without access to subsequent ground-truth
frames.

Let \(\mathcal{E}^{(s,h)}\) denote the corresponding dataset-specific error at
step \(h\), namely VRMSE for TRL2D and GSRD and normalized MSE for REALM. We
report the one-step error
\[
    \mathcal{E}_{\rm one}
    =
    \frac{1}{S}
    \sum_{s=1}^{S}
    \mathcal{E}^{(s,1)},
\]
the final-step error
\[
    \mathcal{E}_{\rm final}
    =
    \frac{1}{S}
    \sum_{s=1}^{S}
    \mathcal{E}^{(s,H)},
\]
The final-step error measures the accumulated error at the longest prediction horizon.

\paragraph{Relative \(L_2\) error.}
For F1-Airfoil Dynamics, NASA-CRM, and DrivAerML, we evaluate point-wise field
accuracy using relative \(L_2\) error. Given a prediction
\(\widehat{\mathbf{Y}}^{(s)}\in\mathbb{R}^{N_s\times d_{\rm out}}\) and its
ground truth \(\mathbf{Y}^{(s)}\), the error is
\[
    \mathcal{E}_{\rm rel}
    =
    \frac{1}{S}
    \sum_{s=1}^{S}
    \frac{
        \left\|
            \widehat{\mathbf{Y}}^{(s)}
            -
            \mathbf{Y}^{(s)}
        \right\|_{F}
    }{
        \left\|
            \mathbf{Y}^{(s)}
        \right\|_{F}
    }.
\]
The values in our tables are reported as percentages. For F1-Airfoil Dynamics,
the five future velocity frames are predicted jointly and concatenated along
the output dimension when computing the error; no autoregressive rollout is
performed. On NASA-CRM, relative \(L_2\) errors are reported separately for
surface pressure \(p_s\) and skin-friction coefficient \(C_f\). On DrivAerML,
we report separate errors for surface pressure \(p_s\), volume velocity
\(\mathbf{u}\), wall shear stress \(\boldsymbol{\tau}\), and volume pressure
\(p_v\).

\paragraph{Integrated aerodynamic coefficients.}
Following Transolver-3, we derive the drag and lift coefficients on DrivAerML
by integrating the predicted surface pressure and wall shear stress over the
full-resolution vehicle surface. The aerodynamic force is
\[
    \mathbf{F}
    =
    \int_{\mathcal{S}}
    \left[
        -
        \bigl(
            p(\mathbf{x})-p_{\infty}
        \bigr)
        \mathbf{n}(\mathbf{x})
        +
        \boldsymbol{\tau}_{w}(\mathbf{x})
    \right]
    \mathrm{d}S,
\]
where \(p_{\infty}\) is the freestream pressure,
\(\mathbf{n}\) is the outward surface normal, and
\(\boldsymbol{\tau}_{w}\) is the wall shear-stress vector. On a discrete
surface mesh, this integral is evaluated as
\[
    \mathbf{F}
    \approx
    \sum_{i=1}^{N_{\mathcal{S}}}
    \left[
        -
        \bigl(
            p_i-p_{\infty}
        \bigr)
        \mathbf{n}_i
        +
        \boldsymbol{\tau}_{w,i}
    \right]
    A_i,
\]
where \(A_i\) is the area associated with surface element \(i\).

Let
\[
    q_{\infty}
    =
    \frac{1}{2}
    \rho_{\infty}
    U_{\infty}^{2}
\]
denote the freestream dynamic pressure and \(A_{\rm ref}\) the reference area.
The drag and lift coefficients are computed as
\[
    C_d
    =
    \frac{
        \mathbf{F}\cdot\mathbf{e}_{d}
    }{
        q_{\infty}A_{\rm ref}
    },
    \qquad
    C_l
    =
    \frac{
        \mathbf{F}\cdot\mathbf{e}_{l}
    }{
        q_{\infty}A_{\rm ref}
    },
\]
where \(\mathbf{e}_{d}\) and \(\mathbf{e}_{l}\) denote the drag and lift
directions, respectively. The same surface quadrature is applied to the
predicted and ground-truth fields.

\paragraph{\(R^2\) score.}
We quantify the agreement between the predicted and reference aerodynamic
coefficients using the coefficient of determination. For
\(C\in\{C_d,C_l\}\),
\[
    R^{2}(C)
    =
    1
    -
    \frac{
        \sum_{s=1}^{S}
        \left(
            \widehat{C}^{(s)}
            -
            C^{(s)}
        \right)^2
    }{
        \sum_{s=1}^{S}
        \left(
            C^{(s)}
            -
            \overline{C}
        \right)^2
    },
    \qquad
    \overline{C}
    =
    \frac{1}{S}
    \sum_{s=1}^{S}
    C^{(s)}.
\]
An \(R^2\) value of \(1\) indicates perfect agreement, \(R^2=0\) corresponds
to predicting the test-set mean, and a negative value indicates performance
worse than this mean predictor.
\subsection{Baselines and Implementation Details}
\label{app:baselines}

We compare Physics Transformer with representative neural operators covering spectral,
latent-token, geometry-aware, and physics-attention architectures. The spectral
baselines include FNO~\citep{li2021fourier}, factorized FNO
(FFNO)~\citep{tran2023factorized}, and GINO~\citep{li2023geometry}. We also
consider latent-token and geometry-aware operators, including
UPT~\citep{alkin2024universal}, AB-UPT~\citep{alkin2025abupt}, and
GAOT~\citep{wen2025geometry}. Physics-attention-based baselines include
Transolver~\citep{wu2024transolver}, LinearNO~\citep{hu2025linearno},
Transolver-3~\citep{wu2026transolver3}, and
GeoTransolver~\citep{adams2025geotransolver}. Together, these baselines cover
regular-grid operators, arbitrary-query latent models, geometry-aware
Transformers, and global physics-slice methods.

\paragraph{FNO.}
The Fourier Neural Operator (FNO)~\citep{li2021fourier} parameterizes global
integral kernels using truncated Fourier modes. Each layer alternates spectral
convolution in the frequency domain with point-wise channel mixing, providing
an efficient operator-learning baseline for physical fields represented on
regular Cartesian grids. We test it on 2D regular case TRL2D and GSRD.

\paragraph{FFNO.}
The Factorized Fourier Neural Operator (FFNO)~\citep{tran2023factorized}
factorizes multidimensional spectral convolution into separate operations along
individual spatial dimensions. Together with its enhanced residual design, this
factorization reduces the cost of spectral mixing and enables deeper neural
operator architectures. FFNO could extend to irregular cases so we test it on 2D regular case TRL2D and GSRD, as well as 2D irregular REALM rocket.

\paragraph{GINO.}
The Geometry-Informed Neural Operator (GINO)~\citep{li2023geometry}
combines graph neural operators with Fourier operator layers to process
physical fields on arbitrary geometries. It encodes irregularly sampled inputs
onto a regular latent grid, performs global operator learning in the Fourier
domain, and decodes the latent representation to arbitrary query locations. We test it on 3D geometry related F1-Airfoil, NASA-CRM and DrivAerML tasks.

\paragraph{UPT.}
The Universal Physics Transformer (UPT)~\citep{alkin2024universal} encodes
variable-size simulation inputs into a fixed-size latent representation through
hierarchical aggregation and latent pooling. A Transformer processes the
compressed latent tokens, and a coordinate-conditioned decoder predicts
physical quantities at the requested output locations. We test it on all tasks.

\paragraph{AB-UPT.}
AB-UPT~\citep{alkin2025abupt} extends UPT to large-scale industrial CFD through
an anchored and branched architecture. It processes geometry, surface fields,
and volumetric fields using dedicated branches and employs anchored latent
representations to support efficient decoding on high-resolution simulation
domains. For it is designed mainly for CFD simulation we test it on F1-Airfoil, NASA-CRM and DrivAerML tasks.

\paragraph{GAOT.}
The Geometry-Aware Operator Transformer (GAOT)~\citep{wen2025geometry}
combines graph-based operator encoders and decoders with a Transformer
processor. Its multiscale graph construction and geometry embeddings enable
information exchange over irregular discretizations while explicitly
incorporating the underlying domain geometry. We test it on all tasks.

\paragraph{Transolver.}
Transolver~\citep{wu2024transolver} introduces Physics-Attention, which
softly assigns discretization points to a small collection of learned physics
slices. Self-attention is performed among these slice tokens, and the updated
representations are subsequently de-sliced to recover point-wise physical
fields. We test it on all tasks.

\paragraph{LinearNO.}
LinearNO~\citep{hu2025linearno} reformulates the slice and de-slice operations
of Physics-Attention from the perspective of linear attention. Through an
algebraic reordering of the associated projections, it obtains a streamlined
neural operator with reduced parameter and computational costs. We test it on 2D regular case TRL2D and GSRD, as well as 2D irregular REALM rocket.

\paragraph{Transolver-3.}
Transolver-3~\citep{wu2026transolver3} scales the Transolver framework to
industrial geometries containing extremely large numbers of cells. It combines
efficient slice computation, geometry slice tiling, subset-based amortized
training, and cached physical states to reduce the memory and computation
required by large-scale training and inference. For Transolver3 does not change the model structure of Transolver, providing mainly training and inference updates, we test it only when Transolver could not handle the whole mesh, on NASA-CRM and DrivAerML.

\paragraph{GeoTransolver.}
GeoTransolver~\citep{adams2025geotransolver} augments the Transolver backbone
with multiscale geometry-aware context. It couples learned physics slices with
geometric and boundary-condition features constructed from local neighborhoods,
allowing the attention processor to incorporate explicit geometric information
on irregular computational domains. For it is designed mainly for CFD simulation we test it on F1-Airfoil, NASA-CRM and DrivAerML tasks.

Unless explicitly identified as literature-reported results, all baselines are
trained and evaluated using the same dataset partitions, input and output
windows, preprocessing, dataset-specific training objectives, optimization
schedule, and evaluation protocol as Physics Transformer. In particular, all methods
use three input frames for one-step prediction on TRL2D, GSRD, and REALM and are
evaluated using the same autoregressive rollout procedure. For
F1-Airfoil Dynamics, all five future frames are predicted jointly under the
official benchmark protocol. On irregular meshes, reproduced methods are given
the same sampled input and query points whenever their architectures permit.
Model widths and depths are adjusted within their original architectural designs
so that all reproduced baselines have parameter counts comparable to
Physics Transformer. We select checkpoints according to validation performance and
apply the same test-time metric implementation to every reproduced method. For relatively
small-scale datasets all models are trained with the full mesh except for larger-scale benchmarks including NASA-CRM and DrivAerML where a subset of mesh (100k) is used to enable training of each model.

We adopt the results on NASA-CRM and DrivAerML from the paper of Transolver-3 and AB-UPT.
~\citep{wu2026transolver3,alkin2025abupt} The GeoTransolver paper reports relative \(L_1\) errors for its point-wise physical-field predictions, whereas our industrial CFD experiments use relative \(L_2\) error. So we reproduce GeoTransolver using its official PhysicsNeMo implementation and evaluate its predictions with the same relative \(L_2\)
implementation used for all other models. For each baseline model, we employed the
AdamW optimizer except for AB-UPT/UPT where LION was used and GeoTransolver where Muon was used.

Training was conducted in either float16 or bfloat16 precision, depending on which yielded better results. All models were trained for equal epochs as shown in table, a weight decay of 1e-4, and gradient clipping at 1. Expect for Transolver3, which is trained for 800 epochs and 600 epochs on DriverML surface and volume separately in line with its released code on GitHub. A cosine learning rate schedule was employed, including a \(5\%\) warm-up phase and a minimum learning rate of 1e-5.
\begin{table*}[t]
    \centering
    \caption{
        Training and model configurations of Physics Transformer.
        All reproduced baselines use the same training configuration for each
        benchmark and are scaled to comparable parameter counts.
        ``Subset Size'' denotes the number of points or cells sampled from each
        instance during training, while ``Full Input'' indicates that no spatial
        subsampling is applied.
        \(L\), \(H\), and \(C\) denote the numbers of layers, attention heads,
        and hidden channels, respectively.
        \(P\) is the number of spatial patches and \(M\) is the number of
        physics tokens per patch, resulting in \(PM\) physics tokens per layer.
    }
    \label{tab:training_model_config}
    \footnotesize
    \setlength{\tabcolsep}{3.2pt}
    \renewcommand{\arraystretch}{1.10}
    \begin{adjustbox}{max width=\textwidth}
    \begin{tabular}{l|ccccc|ccccc}
        \toprule
        &
        \multicolumn{5}{c|}{\textsc{Training Configuration}}
        &
        \multicolumn{5}{c}{\textsc{Model Configuration}}
        \\
        \cmidrule(lr){2-6}
        \cmidrule(lr){7-11}
        \textsc{Benchmark}
        &
        \textsc{Loss}
        &
        \textsc{Epochs}
        &
        \textsc{Initial LR}
        &
        \textsc{Optimizer}
        &
        \textsc{Subset Size}
        &
        \textsc{Layers \(L\)}
        &
        \textsc{Heads \(H\)}
        &
        \textsc{Channels \(C\)}
        &
        \textsc{Tokens \(M\)}
        &
        \textsc{Patches \(P\)}
        \\
        \midrule

        TRL2D
        & \multirow{2}{*}{VRMSE}
        & \multirow{3}{*}{200}
        & \multirow{3}{*}{\(5\times10^{-4}\)}
        & \multirow{3}{*}{AdamW}
        & \multirow{3}{*}{Full Input}
        & \multirow{3}{*}{6}
        & \multirow{3}{*}{4}
        & \multirow{3}{*}{128}
        & \multirow{3}{*}{32}
        & \multirow{3}{*}{512}
        \\
        GSRD
        & & & & & & & & & &
        \\
        REALM
        & MSE & & & & & & & & &
        \\

        \midrule

        F1-Airfoil
        & Relative \(L_2\)
        & 200
        & \(10^{-3}\)
        & AdamW
        & Full Input
        & 6
        & 4
        & 128
        & 32
        & 1024
        \\

        \midrule

        NASA-CRM
        & \multirow{2}{*}{Relative \(L_2\)}
        & \multirow{2}{*}{500}
        & \multirow{2}{*}{\(10^{-3}\)}
        & \multirow{2}{*}{AdamW}
        & \multirow{2}{*}{\(100\mathrm{K}\)}
        & \multirow{2}{*}{16}
        & \multirow{2}{*}{6}& \multirow{2}{*}{192}& \multirow{2}{*}{64}
        & \multirow{2}{*}{1024}
        \\
        DrivAerML
        & & & & & & & & & &
        \\

        \bottomrule
    \end{tabular}
    \end{adjustbox}
\end{table*}

\subsection{Training and Inference Efficiency}
\label{app:efficiency_comparison}

We compare the computational efficiency of \methodname{Physics Transformer} with Transolver-3~\citep{wu2026transolver3} on industry-scale dataset DrivAerML. For the implementation of  Transolver-3 in released code  is different from the configuration reported in its paper, we experiment following the their code on GitHub for training and inference. Training is conducted in parallel on four NVIDIA H800 GPUs with a batch size of \(1\) using 100k points, checkpoint-blocks set to be true. We report the number of trainable parameters,
end-to-end training time, and peak GPU memory allocated on each device.

\begin{table}[t]
    \centering
    \caption{
        Training efficiency compared with Transolver-3. Both models are trained
        on four NVIDIA H800 GPUs with a batch size of \(1\).
        Training time denotes the total wall-clock time, and memory denotes the peak allocation per GPU.
    }
    \label{tab:training_efficiency}
    \tableformat
    \begin{adjustbox}{max width=\linewidth}
    \begin{tabular}{llccc}
        \toprule
        \textsc{Task}
        & \textsc{Model}
        & \textsc{Params. (M)}
        & \textsc{Training Time (h)}
        & \textsc{Memory (GB/GPU)} \\
        \midrule
        \multirow{2}{*}{Surface}
        & Transolver-3
        & 11.33& ~23& 8.3\\
        & Physics Transformer
        & 12.02& ~8& 2.6\\
        \midrule
        \multirow{2}{*}{Volume}
        & Transolver-3
        & 7.6& ~11& 6.5\\
        & Physics Transformer
        & 12.02& ~8& 2.6\\
        \bottomrule
    \end{tabular}
    \end{adjustbox}
\end{table}

We further compare full-resolution inference efficiency. For each surface or
volume mesh, we report the number of queried mesh points, the wall-clock
inference time per instance, and the peak inference memory. Full inference of Transolver-3 follows its tiled inference to obtain the same accuracy performance reported in its papar. While Physics Transformer infer with our proposed token caching and arbitrary query technique. Our method uses bf16, in contrast,  Transolver-3 must use fp32 to ensure precision.

\begin{table}[t]
    \centering
    \caption{
        Full-resolution inference efficiency compared with Transolver-3.
        Inference time is measured per instance after model warm-up, and memory
        denotes peak GPU allocation. The surface and volume meshes are evaluated
        at their respective native resolutions.
    }
    \label{tab:inference_efficiency}
    \tableformat
    \begin{adjustbox}{max width=\linewidth}
    \begin{tabular}{llccc}
        \toprule
        \textsc{Task}
        & \textsc{Model}
        & \textsc{Mesh Points}
        & \textsc{Inference Time (s)}& \textsc{Memory (GB)} \\
        \midrule
        \multirow{2}{*}{Surface}
        & Transolver-3
        & \multirow{2}{*}{\(\sim8\)M}& 168& 5.54\\
        & Physics Transformer
        & & 11& 3.91\\
        \midrule
        \multirow{2}{*}{Volume}
        & Transolver-3
        & \multirow{2}{*}{\(\sim15\)M}& 1524& 4.84\\
        & Physics Transformer
        & & 203& 3.92\\
        \bottomrule
    \end{tabular}
    \end{adjustbox}
\end{table}

\paragraph{Why full-mesh inference is necessary.}
As reported for AB-UPT~\citep{alkin2025abupt} and
Transolver-3~\citep{wu2026transolver3}, increasing the number of mesh points
used during inference can substantially improve the accuracy of the predicted
drag and lift coefficients, \(C_D\) and \(C_L\). We observe the same phenomenon
for Physics Transformer. This capability is important because integrated aerodynamic
coefficients, rather than point-wise fields alone, are often the primary
quantities of interest in industrial CFD. The improvement can be understood
from two aspects. First, a denser input discretization may reduce point-wise
field prediction errors by providing more complete geometric and physical
information. However, our experiments show that the resulting improvement in
point-wise accuracy is relatively small. More importantly, \(C_D\) and \(C_L\)
are computed by numerically integrating the predicted pressure and wall-shear
stress over the vehicle surface:
\[
    \mathbf{F}
    \approx
    \sum_{i=1}^{N}
    \left(
        -p_i\mathbf{n}_i+\boldsymbol{\tau}_i
    \right)A_i,
\]
where \(A_i\) and \(\mathbf{n}_i\) denote the area and normal of the
corresponding surface element. Consequently, errors in the surface quadrature,
particularly in the estimated element areas, directly affect the resulting
aerodynamic coefficients. A denser mesh provides a more accurate representation
of the surface geometry and its integration elements. Moreover, industrial CFD
meshes typically contain precomputed element areas and normals. Performing
prediction on the original full-resolution mesh therefore preserves these
quadrature quantities and enables accurate evaluation of \(C_D\) and \(C_L\),
making efficient whole-mesh inference essential for practical aerodynamic
assessment.
\paragraph{Physics Transformer achieves preferable efficiency.}
As shown in Tables~\ref{tab:training_efficiency}
and~\ref{tab:inference_efficiency}, Physics Transformer is more efficient than
Transolver-3 in both training and inference under broadly comparable parameter
budgets. Transolver-3 adopts a smaller architecture for volumetric prediction,
whereas Physics Transformer uses the same model configuration for both surface and
volume fields. During training, the two models require approximately \(1\)
second per optimization step, and the reduction in total training time achieved
by Physics Transformer primarily results from requiring fewer training epochs. The
most substantial advantage arises during full-resolution inference.
Transolver-3 relies on computationally intensive tiled inference over the entire
mesh to recover its reported accuracy. In contrast, the local physics tokens of
Physics Transformer efficiently represent the physical field using only a relatively
small support set, after which predictions on the complete mesh can be decoded
directly from the cached tokens without accuracy degradation. Consequently,
Physics Transformer achieves approximately \(15.3\times\) and \(7.5\times\) faster
inference on the full surface and volume meshes, respectively. Specifically, it
reduces the per-instance inference time from \(2.8\) minutes to \(11\) seconds
for surface fields and from \(25.4\) minutes to \(3.4\) minutes for volumetric
fields, substantially accelerating the evaluation of aerodynamic performance
on industrial-scale meshes.
\newpage
\section{More visualizations}
\begin{figure}[h]
    \centering
    \includegraphics[width=1\linewidth]{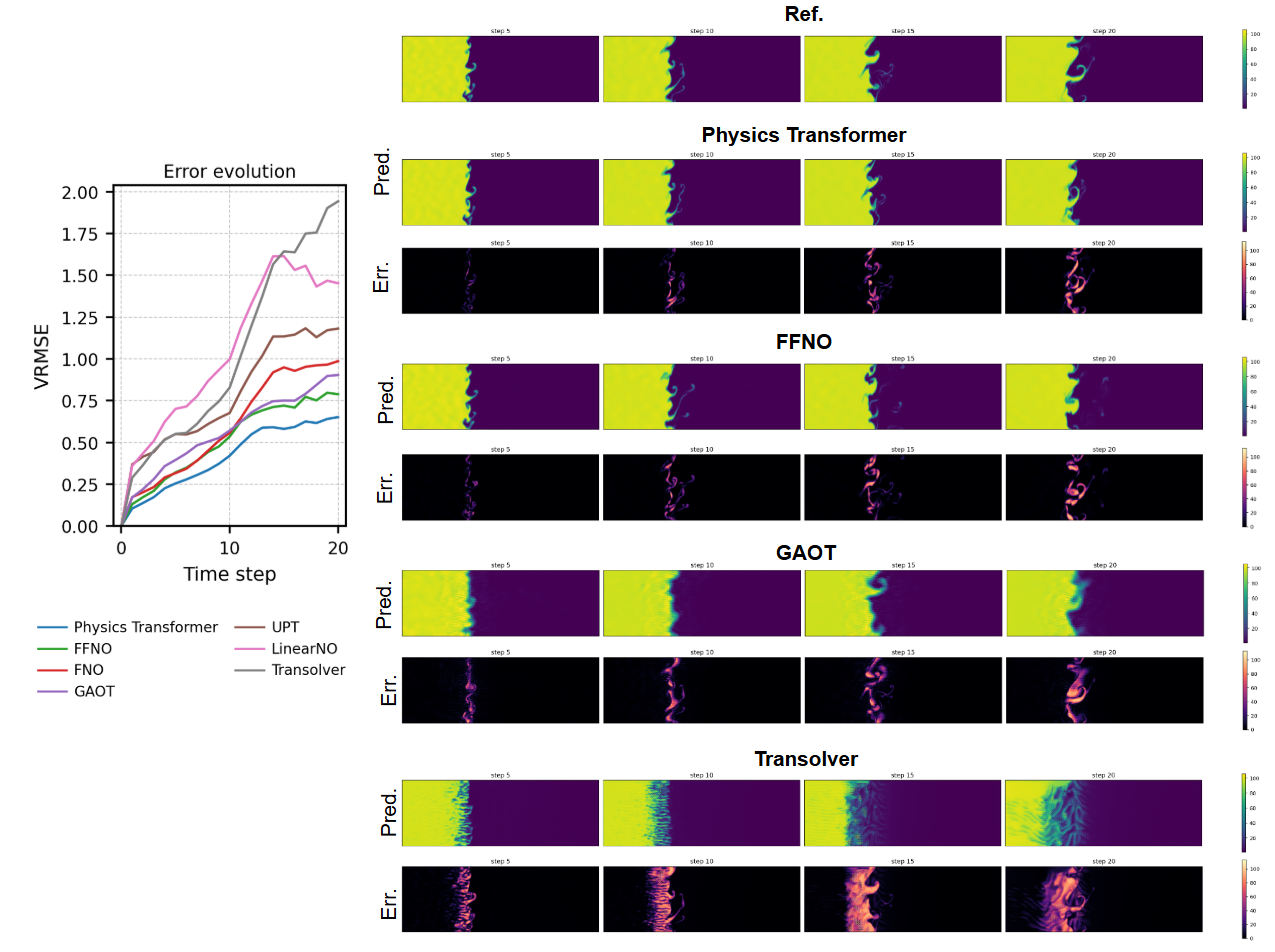}
    \caption{Results on Trl2D dataset.}
    \label{fig:trl}
\end{figure}

\begin{figure}[!t]
    \centering
    \includegraphics[width=1\linewidth]{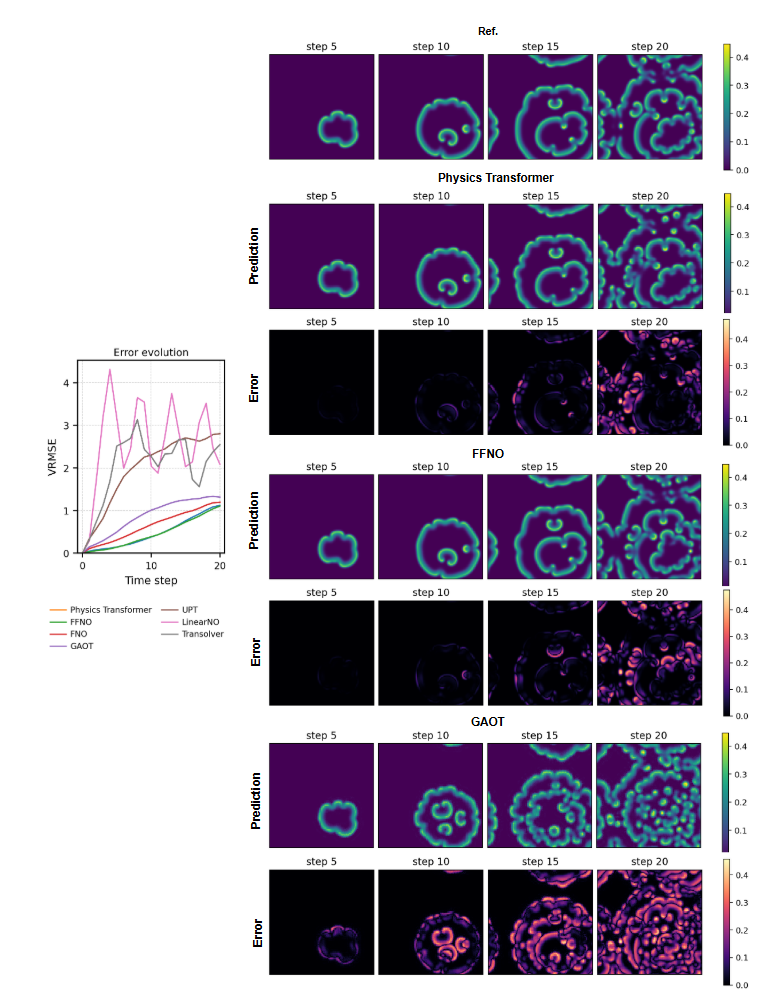}
    \caption{Results on GSDR dataset.}
    \label{fig:gs}
\end{figure}

\begin{figure}[!t]
    \centering
    \includegraphics[width=1\linewidth]{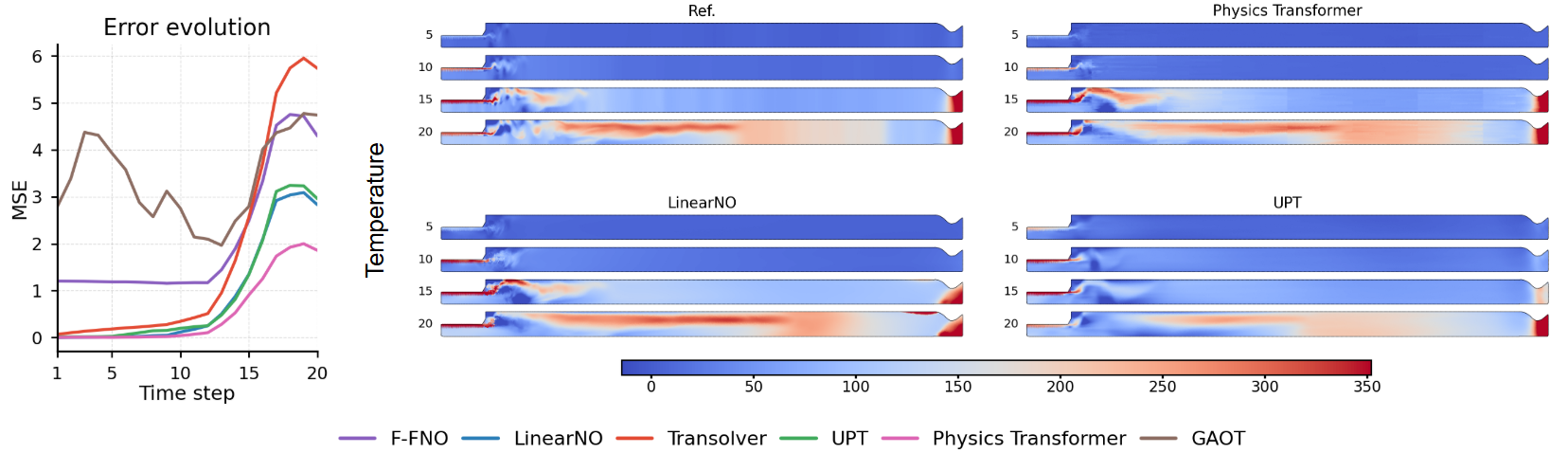}
    \caption{Results on REALM dataset.}
    \label{fig:realm}
\end{figure}

\begin{figure}[!t]
    \centering
    \includegraphics[width=1\linewidth]{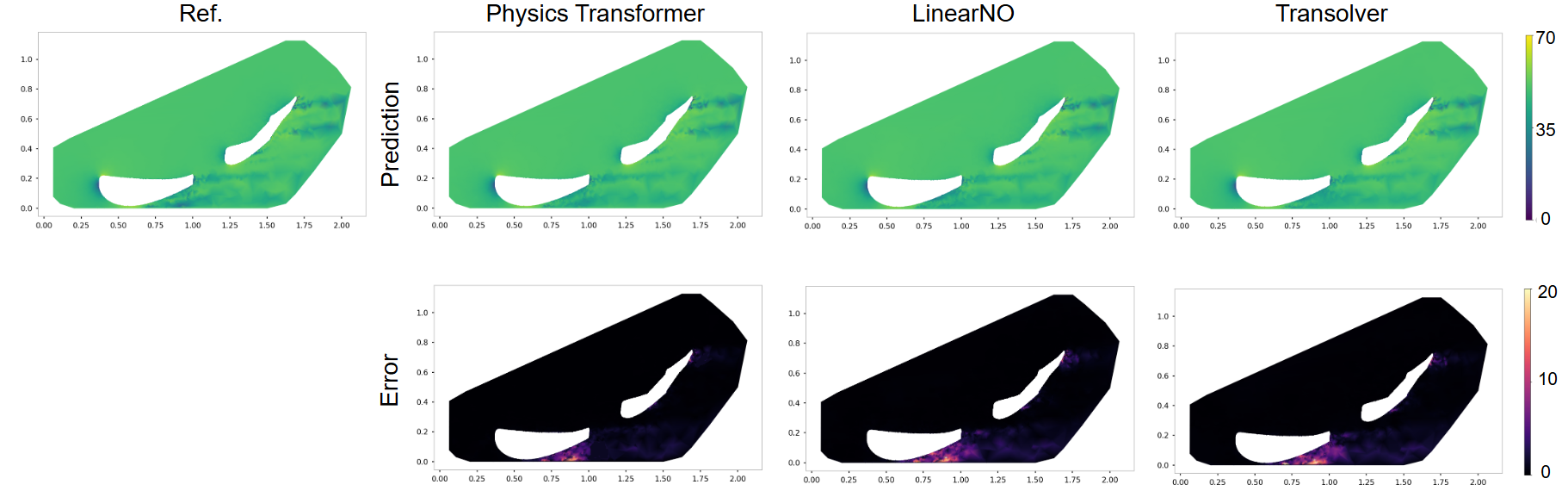}
    \caption{Results on F1-Airfoil Dynamics dataset.}
    \label{fig:gram}
\end{figure}

\begin{figure}[!t]
    \centering
    \includegraphics[width=1\linewidth]{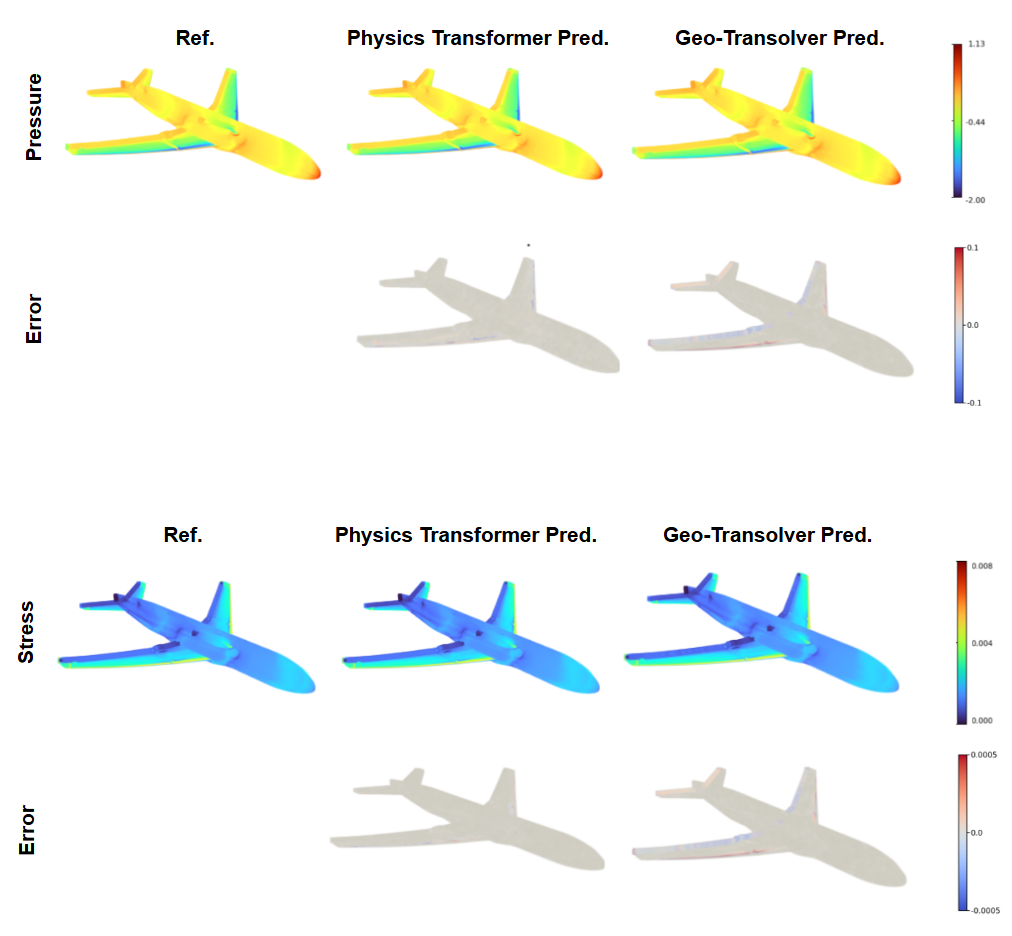}
    \caption{Full results on NASA-CRM dataset.}
    \label{fig:nasa}
\end{figure}

\begin{figure}[!t]
    \centering
    \includegraphics[width=1\linewidth]{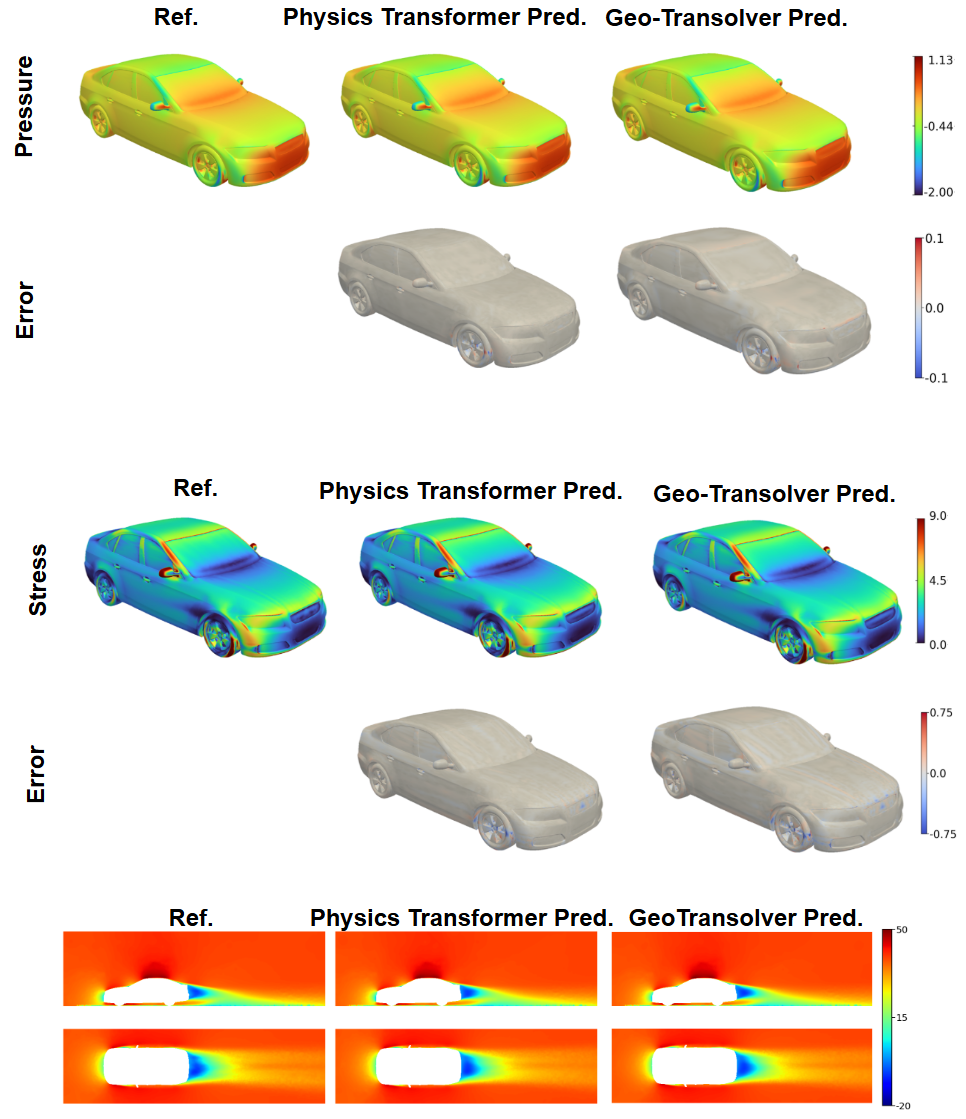}
    \caption{Full results on DrivAerML dataset.}
    \label{fig:car}
\end{figure}

\end{document}

%% file: math_commands.tex
\usepackage{amsmath,amsfonts,bm}

\def\eqref#1{equation~\ref{#1}}
\def\1{\bm{1}}

\DeclareMathAlphabet{\mathsfit}{\encodingdefault}{\sfdefault}{m}{sl}
\SetMathAlphabet{\mathsfit}{bold}{\encodingdefault}{\sfdefault}{bx}{n}